\documentclass{article}

\usepackage{iclr2027_conference,times}

\iclrfinalcopy %

\usepackage[utf8]{inputenc}
\usepackage[T1]{fontenc}
\usepackage{hyperref}
\usepackage{url}
\usepackage{booktabs}
\usepackage{amsfonts}
\usepackage{nicefrac}
\usepackage[table]{xcolor}
\definecolor{idgray}{gray}{0.9} %
\definecolor{redrow}{rgb}{1,0.80,0.80} %
\definecolor{purplerow}{rgb}{0.8,0.8,1} %
\usepackage{graphicx}
\usepackage{capt-of} %
\usepackage{arydshln}
\newcommand{\dashmid}{\noalign{\vskip\aboverulesep}\hdashline\noalign{\vskip\belowrulesep}}
\usepackage{amsmath}
\usepackage{amssymb,amsthm,mathtools,bm}
\usepackage{enumitem}
\usepackage{tikz}
\usetikzlibrary{arrows.meta,calc,positioning,patterns,decorations.pathreplacing,shapes.geometric,fit,backgrounds}

\title{Length Generalization Needs Proper \\ Regularization}

\author{
 \textbf{Pavlo Vasylenko\textsuperscript{1,2}},
 \textbf{Matthias Lindemann\textsuperscript{2}},
 \textbf{André F. T. Martins\textsuperscript{1,2,3}},
 \textbf{Marcos Treviso\textsuperscript{1,2,3,4}}
\\
 \textsuperscript{1}Instituto Superior Técnico, Universidade de Lisboa,
 \textsuperscript{2}Instituto de Telecomunicações, \\
 \textsuperscript{3}ELLIS Unit Lisbon,
 \textsuperscript{4}Gandara AI
}

\newcommand{\R}{\mathbb{R}}
\newcommand{\E}{\mathbb{E}}
\newcommand{\Var}{\mathrm{Var}}
\newcommand{\Cov}{\mathrm{Cov}}

\newcommand{\RMSNorm}{\mathrm{RMSNorm}}

\newcommand{\vv}[1]{\bm{#1}}
\newcommand{\mM}[1]{\bm{#1}}

\newtheorem{proposition}{Proposition}

\newcommand{\RMS}{\mathrm{RMS}}

\begin{document}

\maketitle
\lhead{Preprint} %

\begin{abstract}

Length generalization is the ability of sequential models to perform well on context lengths unseen during training.
In this work, we show that the challenge of achieving length generalization is related not only to architectural choices such as positional encoding and the attention mechanism but also to the training procedure itself.
We study how regularization affects length generalization and find that weight decay can hinder extrapolation.
In contrast, dropout improves extrapolation when its placement within the architecture is reconsidered.
In that regard, we show that the standard placement of dropout before layer normalization introduces a systematic distributional mismatch, and that applying dropout just before the linear projection resolves this issue.
For example, a modified SmolLM3 with sliding window attention, continually pre-trained with dropout, can extrapolate perfectly to 64$\times$ on Needle-in-a-Haystack and far beyond the pre-training context size on RULER and HELMET. Mamba2 also benefits from dropout, suggesting an architecture-agnostic nature of the problem.
We further propose Variance-Preserving Affine Dropout (VPAD), a new dropout strategy that substantially reduces the resulting pre-activation variance mismatch, leading to further extrapolation improvements in transformers.
\footnote{Code: https://github.com/deep-spin/len-gen-regularization}

\end{abstract}

\section{Introduction}
As we demand that large language models (LLMs) operate over increasingly longer contexts, length generalization remains a central challenge for long-context modeling.
To avoid the challenges of length generalization, the community relies not on the extrapolation ability of models but on continued training on longer contexts after pre-training on much shorter ones~\citep{fu2024data, gao-etal-2025-train, Basant2025NVIDIANN, nvidia2026nemotron3ultraopen, olmo2026olmo3}.
The phenomenon of poor length extrapolation is ubiquitous across the most widely used architectures, such as transformers and linear RNNs~\citep{pitorro2024effective, buitrago2025understanding, ben-kish2025decimamba, Jelassi2024RepeatAM}.
For transformers, key obstacles leading to poor out-of-distribution (OOD) performance are the Positional Encoding (PE)~\citep{press2022train, kazemnejad2023impact, ruoss2023randomized, su2024roformer} and the attention mechanism itself~\citep{velickovic2024softmax, barbero2024transformers, vasylenko2026longcontext}.
For linear RNNs, these challenges have been linked to their fixed-size memory mechanism~\citep{Jelassi2024RepeatAM, arora2024zoology} and, as recently observed by \citet{buitrago2025understanding}, to a train-test mismatch in the distribution of recurrent states.

In this work, we show that the problem of length generalization is more than an architectural challenge and can be related to overfitting to the in-distribution (ID) length, i.e., the pre-training context length. \citet{buitrago2025understanding} showed that training linear RNNs for longer worsens length extrapolation.
We demonstrate that longer training degrades length extrapolation not only in linear RNNs but in transformers as well. Moreover, we show that, despite being a standard regularization technique, weight decay (WD) significantly speeds up the deterioration. This eventually leads to a complete loss of extrapolation. In contrast, regular dropout \citep{srivastava2014dropout} offers significant benefits,
but its standard placement in common implementations leads to a mismatch of a neuron's pre-activation statistics between training and inference: normalization applied to the masked input attenuates the mean of the signal reaching the linear projection.
We show that dropout must instead be placed just before the linear projection, so that normalization acts on a non-masked input.
Additionally, because regular dropout still leads to a variance mismatch, we derive Variance-Preserving Affine Dropout (VPAD)---a generalization of Alpha-Dropout~\citep{klambauer2017selfnormalizing} to non-centered activations---which also preserves the variance of the input to the linear projection.
We show that both dropout variants substantially improve the extrapolation of transformers. For instance, the SmolLM3 checkpoint pre-trained on 4K context length and continually pre-trained on the same length with dropout extrapolates perfectly to $64\times$ on Needle-in-a-Haystack and far beyond its pre-training context on RULER and HELMET. Similarly, Mamba2 benefits, improving its extrapolation by at least $4\times$ on RULER and HELMET.
Overall, we show that \emph{length-based overfitting} affects multiple common architectures, and proper regularization can substantially mitigate it.

\section{Longer Training Hinders Length Generalization}\label{sec:long-train-hinder-lengen}

\begin{figure}
\centering
 \includegraphics[width=0.44\linewidth]{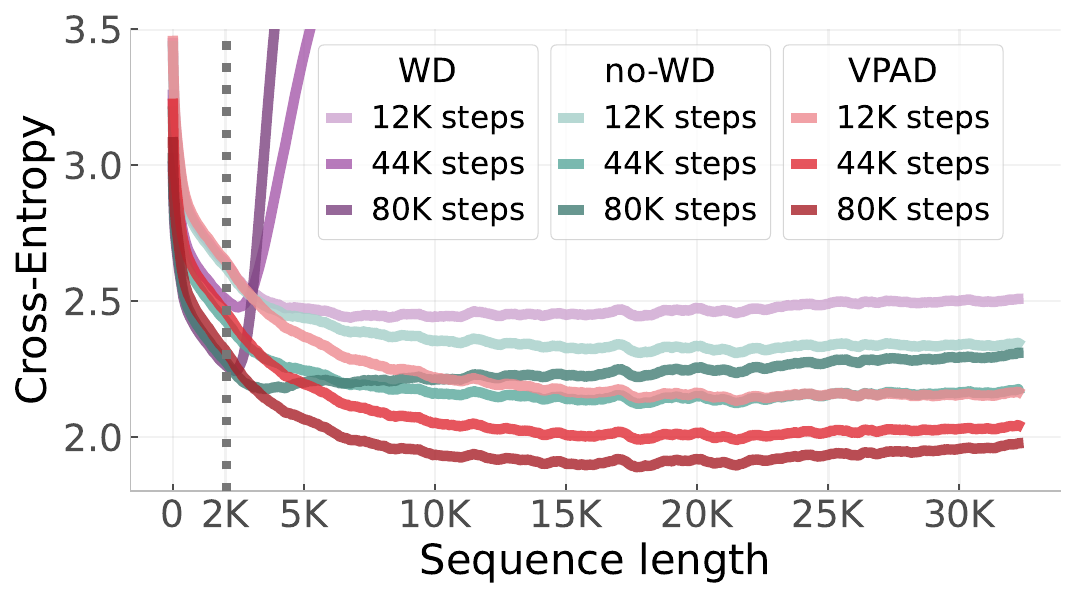}
 \includegraphics[width=0.44\linewidth]{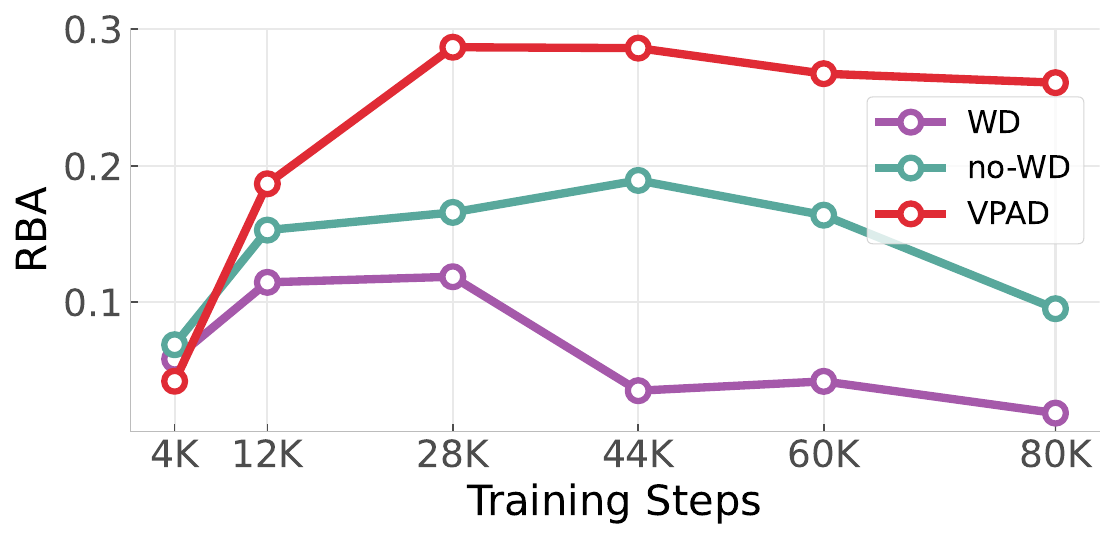}
  \caption{Transformer performance across training steps. \textbf{Left}: per-position CE loss (512-window moving average); the dotted line denotes the pre-training context length (2K). \textbf{Right}: RULER extrapolation, RBA averaged over 4--8K.
  Training with WD leads to overfitting to the pre-training context length. In contrast, dropout (VPAD) substantially mitigates the issue.
  }
 \label{fig:eval-per-step-softmax}
\end{figure}

We hypothesize that the training duration affects length generalization in transformers, mirroring the phenomenon observed for linear RNNs by \citet{buitrago2025understanding}: a model trained on a small number of tokens extrapolates better than one trained for longer.
Fig.~\ref{fig:eval-per-step-softmax} shows that longer training inevitably deteriorates extrapolation.
In particular, the prevalent training approach with weight decay leads to the most severe outcome: in both Cross-Entropy (CE) and Recall-Based Accuracy (RBA) on the RULER benchmark~\citep{hsieh2024ruler}, optimal performance is reached at around 12--28K steps, after which the model degrades.
Interestingly, training without weight decay (no-WD) substantially postpones this effect.
Optimal performance moves to 44K, with a lower and flatter CE loss and a $55\%$ higher RULER score.
Eventually, however, longer training degrades the no-WD model as well: RBA drops to almost half of its maximum by the end of training.

This experiment confirms that transformers are also affected by the issue in \citet{buitrago2025understanding}.
We hypothesize that models overfit to the training context length,
and this can be mitigated by using methods aimed at improving generalization more broadly. In the following sections, we show that dropout substantially alleviates the problem, allowing models to extrapolate much better.

\section{Background}
\label{sec:background}

\paragraph{Transformers and Mamba2 Blocks.} Transformers \citep{vaswani2017attention} and Mamba2 \citep{dao2024mamba2} both process an input sequence by stacking $L$ identical blocks. In transformers, each block is composed of a self-attention sublayer and a position-wise feed-forward network (FFN) sublayer, both wrapped in residual connections together with a normalization layer. In modern transformers such as GPT-2 \citep{radford2019language}, LLaMA \citep{touvron2023llama}, Mistral \citep{jiang2023mistral}, and Qwen \citep{qwen3}, the full block performs the following:
\begin{equation}\label{eq:pre-norm-block}
\mM{X}^{i} \;=\; \mM{R}^{i-1} + \mathrm{Attn}(\mathrm{Norm}(\mM{X}^{i-1})),
\quad
\mM{X}^{i+1} \;=\; \mM{R}^{i} + \mathrm{FFN}(\mathrm{Norm}(\mM{X}^{i})),
\end{equation}
where $\mM{R}^{i}=\mM{X}^{i}$, for easiness of notation, is the residual stream
of sublayer $i \in \{1,\ldots,2L\}$; $\mathrm{Norm}(\cdot)$ is normalization applied row-wise to its input.  Mamba2 follows the same convention, wrapping each block as $\mM{X}^{i} = \mM{R}^{i-1} + \mathrm{Mamba2}(\mathrm{Norm}(\mM{X}^{i-1}))$.

In both architectures, each sublayer first applies a linear projection to its input before any non-linear processing. Given the normalized input $\mM{\hat{X}}=\mathrm{Norm}(\mM{X}) \in \R^{n \times d}$ and input projections $\mM{W}_{\{Q, K, V\}} \in \R^{d \times d}$, the attention sublayer---single-head for brevity\footnote{Our analysis is not affected by the number of heads.}---computes
\begin{equation}
	\mathrm{Attn}: \;
     [\mM{Y}_Q,\mM{Y}_K,\mM{Y}_V] = [\mM{\hat{X}}\mM{W}_Q, \mM{\hat{X}}\mM{W}_K, \mM{\hat{X}}\mM{W}_V],
	\quad
    \mM{U}_{\mathrm{Attn}} =
    \mathrm{softmax}\!\left(s\mM{Y}_Q \mM{Y}_K^\top\right)\mM{Y}_V\,\mM{W}_P,
	\label{eq:attn}
\end{equation}
where $\mM{W}_P \in \R^{d \times d}$ is the output projection; $s$ is an attention scaling factor which is usually set to $d^{-1/2}$ but can be length-dependent \citep{chiang-cholak-2022-overcoming, nakanishi2025scalable, vasylenko2026longcontext}. For the FFN, the input projections are $\mM{W}_{\{G,U\}} \in \R^{d \times d_{\mathrm{ffn}}}$, and the sublayer computes
\begin{equation}
\mathrm{FFN}: \;
      [\mM{Y}_{\mathrm{gate}},\mM{Y}_{\mathrm{up}}] = [\mM{\hat{X}}\mM{W}_G, \mM{\hat{X}}\mM{W}_U],
	\quad
    \mM{U}_{\mathrm{FFN}} =
    \bigl(\phi(\mM{Y}_{\mathrm{gate}}) \odot \mM{Y}_{\mathrm{up}}\bigr)\,\mM{W}_D,
\end{equation}
where $\phi$ is an elementwise nonlinearity and $\mM{W}_D \in \R^{d_{\mathrm{ffn}} \times d}$ is the down projection. In Mamba2, an input projection $\mM{W}_{\mathrm{in}} \in \R^{d \times \tilde{d}}$ produces the gate $\mM{Z}$, the SSM input $\mM{X}_{\mathrm{ssm}}$, the state matrices $\mM{B}$ and $\mM{C}$, and the step size $\vv{\Delta}$. Omitting the short convolution, the output normalization, and the output projection for clarity, and using the SSM-attention duality, the block computes
\begin{equation}
\mM{Y}_{\mathrm{SSD}} = [\mM{Z}, \mM{X}_{\mathrm{ssm}}, \mM{B}, \mM{C}, \vv{\Delta}] \;=\; \mM{\hat{X}}\mM{W}_{\mathrm{in}},
\;
\mathrm{Mamba2}(\mM{Y}_{\mathrm{SSD}})
\;=\; \bigl((\mM{L} \odot
\mM{C}\mM{B}^\top)\,\mM{X}_{\mathrm{ssm}}\bigr) \odot \phi(\mM{Z}),
\label{eq:mamba2}
\end{equation}
where $\mM{L} \in \R^{n \times n}$ is a causal decay mask built from $\vv{\Delta}$ and a learned decay rate.
\paragraph{RMSNorm.} The original choice of $\mathrm{Norm}(\cdot)$ is LayerNorm \citep{ba2016layernorm}, but modern LLMs conventionally replace it with Root Mean Square Normalization \citep{zhang2019rmsnorm}:
\begin{equation}\label{eq:rms-def}
\RMSNorm(\vv{x}) \;=\; \vv{\gamma} \odot \frac{\vv{x}}{\RMS(\vv{x}) + \epsilon}, \qquad \RMS(\vv{x}) \;=\; \sqrt{\frac{1}{d}\sum_{i=1}^d x_i^2}.
\end{equation}

\paragraph{Dropout.}
Dropout \citep{srivastava2014dropout} is a regularization method used in many transformer architectures.
At training time, dropout replaces each element of a hidden vector $\vv{u} \in \R^d$ by
$\mathrm{Drop}(\vv{u})_i = \frac{b_i}{1-p}\, u_i$,
where $b_1, \dots, b_d \sim \mathrm{Bernoulli}(1-p)$ are i.i.d.\ masks and $p \in [0,1)$ is the drop probability. The $1/(1-p)$ rescaling (known as ``inverted dropout'') preserves the expectation of pre-activations during training, $\E[\vv{u}] = \E[\mathrm{Drop}(\vv{u})]$.
At inference time, no masking is applied.
Within a transformer block, dropout is conventionally applied to the output of each sublayer just before the residual addition, so any two successive sublayers execute:
\begin{equation}\label{eq:dropout-transformers}
\tilde{\mM{X}}^{i} \;=\; \mM{R}^{i-1} + \mathrm{Drop}\bigl(\underbrace{\mathrm{Attn}(\mathrm{Norm}(\tilde{\mM{X}}^{i-1}))}_{\mM{U}_\mathrm{Attn}}\bigr),
\quad
\tilde{\mM{X}}^{i+1} \;=\; \mM{R}^{i} + \mathrm{Drop}\bigl(\underbrace{\mathrm{FFN}(\mathrm{Norm}(\tilde{\mM{X}}^{i}))}_{\mM{U}_\mathrm{FFN}}\bigr),
\end{equation}
where $\mathrm{Drop}(\cdot)$ is dropout applied row-wise to its input with an independent mask $\vv{b}^j$ for each row $j$. As a result, the effect of dropout on the next linear layer (e.g. from an attention or feedforward sublayer) is mediated through an RMSNorm (Fig.~\ref{fig:dropout-placements}, Left).
We revisit this placement---Pre-Norm dropout---in Sec.~\ref{sec:theory} and show that it introduces a systematic mismatch between training-time and inference-time pre-activation statistics. As we will show, dropout can be placed just before linear projections to mitigate the problem (Pre-Linear dropout, Sec.~\ref{sec:pl}; see also Fig.~\ref{fig:dropout-placements}, Right):
\begin{equation}\label{eq:dropout-pre-linear}
\mM{X}^{i} \;=\; \mM{R}^{i-1} + \underbrace{\mathrm{Attn}(\mathrm{Drop}(\mathrm{Norm}(\mM{X}^{i-1})))}_{\mM{U}_\mathrm{Attn}},
\quad
\mM{X}^{i+1} \;=\; \mM{R}^{i} + \underbrace{\mathrm{FFN}(\mathrm{Drop}(\mathrm{Norm}(\mM{X}^{i})))}_{\mM{U}_\mathrm{FFN}}.
\end{equation}

\section{Dropout and Train-Test Mismatch}\label{sec:theory}

In common transformer implementations, such as in HuggingFace transformers~\citep{wolf-etal-2020-transformers} and Megatron \citep{megatron-lm}, dropout is typically applied before adding the residual (Eq. \ref{eq:dropout-transformers}), that is, before the normalization layer of the next transformer block.
In Sec.~\ref{sec:pn}, we show that this placement---\emph{Pre-Norm} dropout---introduces a systematic mismatch between the pre-activation means
seen at training and at inference. Sec.~\ref{sec:pl} shows that \emph{Pre-Linear} dropout (Eq.~\ref{eq:dropout-pre-linear}), which masks the normalized representation just before the linear layer, does not suffer from the issue.
Finally, Sec.~\ref{sec:variance-preserve-drop} describes a variance-preserving variant of Pre-Linear dropout that also reduce the variance matches.
Full proofs of the propositions below are given in App.~\ref{app:proofs-rms}.

\subsection{Theoretical Setup}\label{sec:preact-setup}
We first introduce concepts and assumptions used for our theoretical analysis.
 We study the pre-activation of a single-neuron linear layer that follows a normalization layer.
 Let $\vv{x} = (x_1, \dots, x_d) \in \R^d$ be the input to the normalization layer. We assume that $\vv{x}$ is a multivariate random variable with components $x_i$ drawn i.i.d. from a distribution with mean $\mu_x := \E[x_i]$ and variance $\sigma_x^2 := \Var[x_i]$. This is a common assumption \citep{he2015delving, pmlr-v70-gehring17a, vankadara2024on} that simplifies the derivations.

For brevity, let $\hat{x}_i := x_i/(\RMS(\vv{x}) + \epsilon)$ be the output of the RMSNorm transformation with $\bm{\gamma}=\mathbf{1} \in \R^d$.  Let $\vv{w} \in \R^d$ be the weights of a single output neuron, and let $\vv{z}$ denote whatever is actually fed into that neuron---which depends on whether and where dropout is applied. For instance, at inference time, no dropout is applied, so $\vv{z} = \hat{\vv{x}}$ and the \emph{pre-activation} is simply $y^{\mathrm{test}} = \vv{w}^\top \vv{z} = \vv{w}^\top \hat{\vv{x}} \in \R$.%
\footnote{We omit the bias term as it does not affect our analysis: the training- and inference-time expectations are shifted by the same constant, and the variance is translation-invariant.}
As $d\to\infty$, by the law of large numbers, $\RMS(\vv{x}) \to \sqrt{\mu_x^2 + \sigma_x^2}$ (see App.~\ref{app:proofs-baseline}), so the mean $\mu_{\hat{x}} := \E[\hat{x}_i]$ and variance $\sigma_{\hat{x}}^2 := \Var[\hat{x}_i]$ of the normalized coordinate satisfy $\mu_{\hat{x}} \to \frac{\mu_x}{\sqrt{\mu_x^2 + \sigma_x^2} + \epsilon}$ and $\sigma_{\hat{x}}^2 \to \frac{\sigma_x^2}{(\sqrt{\mu_x^2 + \sigma_x^2} + \epsilon)^2}$,
 and the pre-activation has baseline mean and variance:
\begin{equation}
\E[y^{\mathrm{test}}] \;=\; \mu_{\hat{x}} \sum_{i=1}^d w_i, \qquad \Var[y^{\mathrm{test}}] \;=\; \sigma_{\hat{x}}^2\,\|\vv{w}\|^2 + C(\vv{\hat{x}}, \vv{w}),
\label{eq:baseline}
\end{equation}
where $C(\hat{\vv{x}}, \vv{w})=\sum_{i \neq j} w_i w_j\,\Cov[\hat{x}_i, \hat{x}_j]$ and $\Cov[\hat{x}_i, \hat{x}_j]$
is the cross-coordinate covariance induced by the shared normalization denominator (the $\hat{x}_i$ are no longer i.i.d. but exchangeable).
Eq.~\eqref{eq:baseline} is the inference baseline against which Pre-Norm, Pre-Linear, and VPAD are compared.

\begin{figure}
\centering
  \includegraphics[width=0.49\linewidth]{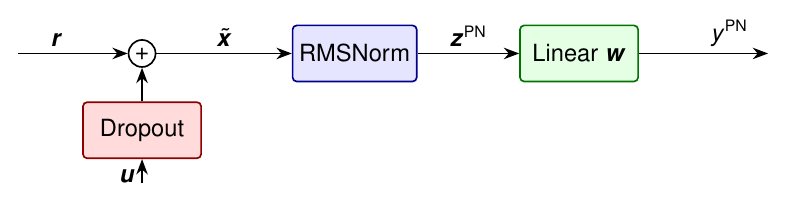}\hfill
  \includegraphics[width=0.49\linewidth]{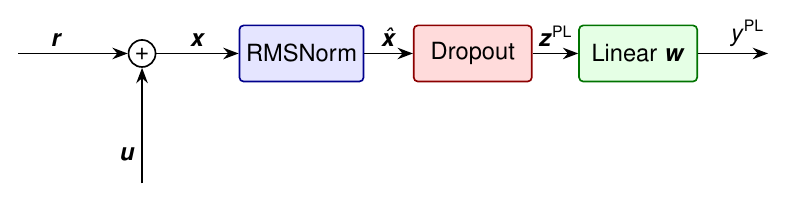}
  \caption{Dropout placements. \textbf{Left}: Pre-Norm dropout (Eq.~\ref{eq:dropout-transformers}). \textbf{Right}: Pre-Linear dropout. $\mM{r}$ --- residual stream, $\mM{u}$ --- output of a sublayer (e.g. $\mathrm{Attn}$, $\mathrm{FFN}$, $\mathrm{Mamba2}$), Linear --- linear projection.
  }
  \label{fig:dropout-placements}
\end{figure}

\subsection{Pre-Norm dropout}\label{sec:pn}

Under Pre-Norm dropout---the conventional placement, Eq.~\ref{eq:dropout-transformers}---the masked sublayer output is added to the residual stream, and the sum is what the next normalization layer reads (Fig.~\ref{fig:dropout-placements}, Left). We therefore split the normalization input into $\vv{x} = \vv{r} + \vv{u}$,
where $\vv{u}$ (mean $\mu_u$, variance $\sigma_u^2$) is the current sublayer output---the component the dropout mask acts on---and $\vv{r}$ is the residual accumulated by earlier sublayers.
At training time, dropout replaces $\vv{u}$ by $\tfrac{\vv{b}}{1-p}\odot\vv{u}$ before the residual addition, so the normalized vector and the quantity fed to the linear layer are
\begin{equation}
\tilde{x}_i \;=\; r_i + \frac{b_i}{1-p}\,u_i, \qquad z_i^{\mathrm{PN}} \;=\; \frac{\tilde{x}_i}{\RMS(\tilde{\vv{x}}) + \epsilon}, \qquad
y^{\mathrm{PN}} = \vv{w}^\top \vv{z}^{\mathrm{PN}}.
\label{eq:pn-input}
\end{equation}

\begin{proposition}[Pre-Norm dropout]\label{prop:pn}
Assume $\mu_x \neq 0$. As $d \to \infty$,
\begin{equation}
\begin{aligned}
\E[y^{\mathrm{PN}}] \;&\longrightarrow\; c_\epsilon\,\E[y^{\mathrm{test}}],
\qquad
c_\epsilon
\;=\; \Bigl(1 + \tfrac{p}{1-p}\,\kappa\Bigr)^{-1/2} + O(\epsilon),
\end{aligned}
\end{equation}
where $\kappa \;=\; \E[u_i^2]/\E[x_i^2]$.
\end{proposition}
Whenever $\mu_x \neq 0$ and $p > 0$, Pre-Norm dropout attenuates the pre-activation mean, creating a train-test mismatch in the inputs to the linear projection layer.
Since the mean shifts, this can result in activation of completely different neurons, disrupting the subsequent layers, which in turn can propagate error to the next tokens by  attention.

\subsection{Pre-Linear dropout}\label{sec:pl}

The alternative placement applies RMSNorm to the residual stream first, and only then standard inverted dropout, immediately before the linear layer (Fig.~\ref{fig:dropout-placements}, Right): $z_i^{\mathrm{PL}} \;=\; \frac{b_i}{1-p}\,\hat{x}_i$.
Because the mask now acts on the already-normalized representation, the result is independent of how $\vv{x}$ decomposes into residual and sublayer contributions.
The pre-activation moments are
\begin{equation}
\E[y^{\mathrm{PL}}] \;=\; \E[y^{\mathrm{test}}],
\qquad
\Var[y^{\mathrm{PL}}] \;=\; \Var[y^{\mathrm{test}}] \;+\; \frac{p}{1-p}\,\E[\hat{x}_i^2]\,\|\vv{w}\|^2.
\label{eq:pl-moments}
\end{equation}
That is, Pre-Linear dropout preserves the mean exactly but inflates the variance, irrespective of the input.
This is the familiar behavior of inverted dropout: the mask preserves expectations, but the extra randomness it introduces is not removed by the $1/(1-p)$ rescaling~\citep{hendrycks2017adjusting, dopout-batch-norm-disharmony}; see the proof in App.~\ref{app:proof-pl} (Proposition~\ref{prop:pl}).

\subsection{Variance-Preserving Affine Dropout}\label{sec:variance-preserve-drop}
Since Pre-Linear dropout inflates the variance for any $p > 0$, we hypothesize that, similar to the train-test mismatch of the mean, the variance mismatch can affect the subsequent layers as well.
We therefore derive \emph{Variance-Preserving Affine Dropout} (VPAD): a deterministic affine correction of the masked activation so that \emph{both} moments are preserved. With mean $\mu_{\hat{x}}$ and variance $\sigma_{\hat{x}}^2$
of the normalized input $\hat{x}_i$ (see Sec.~\ref{sec:preact-setup}), VPAD replaces the masked activation $b_i\hat{x}_i$ by
\begin{equation}
z_i^{\mathrm{VPAD}} \;=\; \alpha\, b_i\, \hat{x}_i \;+\; \beta,
\qquad
\alpha \;=\; \sigma_{\hat{x}}((1-p)\,(\sigma_{\hat{x}}^2 + p\,\mu_{\hat{x}}^2))^{-1/2},
\qquad
\beta \;=\; \mu_{\hat{x}}\bigl(1 - \alpha(1-p)\bigr).
\label{eq:vpad-coef}
\end{equation}

\begin{figure}
    \centering
    \includegraphics[width=0.99\linewidth]{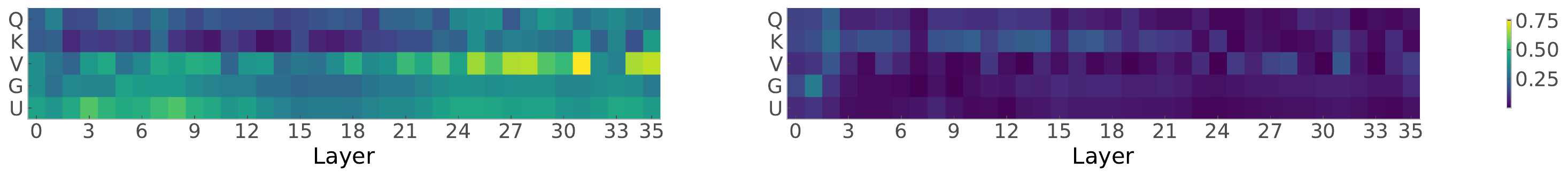}
    \caption{Comparison of the relative error $|\Var[y^{\mathrm{Dropout}}] - \Var[y^{\mathrm{test}}]|/\Var[y^{\mathrm{test}}]$ of the pre-activation variances across layers of SmolLM3 fine-tuned with \textbf{Left}: Pre-Linear Dropout; \textbf{Right}: VPAD. VPAD substantially reduces the error. Moreover, the error decreases with depth for VPAD but grows for PL-D (first/last 18 layers, averaged over matrices and layers: 0.09/0.06 vs.\ 0.29/0.37).}
    \label{fig:diff-variance}
\end{figure}

The coefficients $\alpha$ and $\beta$ are
chosen so $\E[z_i^{\mathrm{VPAD}}] = \mu_{\hat{x}}$ and $\Var[z_i^{\mathrm{VPAD}}] = \sigma_{\hat{x}}^2$. Consequently the pre-activation matches the inference baseline exactly in the mean, and in the variance up to an attenuated covariance term:
\begin{equation}
\E[y^{\mathrm{VPAD}}] \;=\; \E[y^{\mathrm{test}}],
\qquad
\Var[y^{\mathrm{VPAD}}] - \Var[y^{\mathrm{test}}] \;=\; \bigl(\alpha^2(1-p)^2 - 1\bigr)\,C(\vv{\hat{x}}, \vv{w})
\label{eq:vpad-preact}
\end{equation}
(Proposition~\ref{prop:darms}, App.~\ref{app:proof-darms}). The masks attenuate each pairwise covariance by the factor $\alpha^2(1-p)^2 = \tfrac{\sigma_{\hat{x}}^2(1-p)}{\sigma_{\hat{x}}^2 + p\,\mu_{\hat{x}}^2} \in (0,1)$, so VPAD matches the per-coordinate variances exactly and shrinks the covariance term; hence,
the residual variance mismatch never exceeds $|C(\vv{\hat{x}}, \vv{w})|$. In contrast, Pre-Linear dropout has fixed $\tfrac{p}{1-p}\|\vv{w}\|^2$ inflation. Fig.~\ref{fig:diff-variance} compares the relative errors for two SmolLM3 models fine-tuned with either Pre-Linear dropout or VPAD: VPAD substantially reduces the error, and later layers exhibit smaller errors than the earlier ones.

In practice, we replace $\mu_{\hat{x}}$ and $\sigma_{\hat{x}}^2$ by the empirical estimates $\tfrac{1}{d}\sum_i \hat{x}_i$ and $\tfrac{1}{d}\sum_i (\hat{x}_i - \mu_{\hat{x}})^2$, which concentrate around their population values for large $d$. For $\mu_{\hat{x}}=0$ and $\sigma_{\hat{x}}^2=1$, the coefficients~\eqref{eq:vpad-coef} recover the affine correction of Alpha-Dropout~\citep{klambauer2017selfnormalizing} with the dropped value set to zero ($\alpha'=0$): VPAD is thus its extension to non-centered activations---exactly the situation created by RMSNorm. Another distinction from Alpha-Dropout is that $\alpha$ and $\beta$ are computed per token while Alpha-Dropout does not depend on the input at all.

\section{Experiments}\label{sec:experiments}

\subsection{Experimental Setup}\label{sec:experimental-setup}
We conduct experiments in two settings: training from scratch and continual pre-training (CPT). In all experiments, we apply weight decay only to the models labeled ``WD''.
Note that we train the different setups with different amounts of data.
The results must therefore be read within each experimental setting and model architecture. We report all experimental details in App. \ref{appendix:experiment-details}.

\paragraph{Training from scratch} For experiments with transformers, we use a Llama-3-like model but replace RoPE with NAPE \citep{vasylenko2026longcontext}, which has shown its effectiveness at extrapolation;
we use no attention scaling or any other intervention related to PE. We train 400M-parameter models with a 2K context length on 40B tokens in five settings: weight decay (WD), no weight decay (no-WD), Pre-Norm Dropout (PN-D), Pre-Linear Dropout (PL-D), and VPAD.
For Mamba2 \citep{dao2024mamba2}, we follow \citet{buitrago2025understanding} and train 85M-parameter models with a 1K context length on 25B tokens.
For evaluation, we measure CE loss on 2K-long documents
and Recall-Based Accuracy (RBA) on Needle-in-a-Haystack (NIAH-1 from RULER) and overall RULER.

\paragraph{Continual Pre-training}
To test dropout at scale, we employ continual pre-training of larger models. For transformers, we use the SmolLM3-3B \citep{bakouch2025smollm3} checkpoint (stage 3) pre-trained with a 4K context. We convert its RoPE layers to sliding-window (SW) attention  with a window size of 4K while leaving the NoPE layers untouched. Models are then trained with the same 4K context size on 1B tokens. To test the complementarity of dropout and attention scaling, we train models---SmolLM3-SW-8K---with an 8K context and a 4K sliding window, as well as a variant with the input- and length-dependent attention scaling of \citet{vasylenko2026longcontext}, denoted S$\times$. We compare the SmolLM3-SW-8K models with the original SmolLM3-Base, which was trained on an additional 100B tokens after stage 3, and with SmolLM3-Base extended with YaRN RoPE.
For Mamba2, we apply CPT to a 1.3B-parameter model for 0.5B tokens.
We evaluate all models on NIAH-1, RULER \citep{hsieh2024ruler}, and a subset of HELMET \citep{yen2025helmet} tasks: In-Context Learning (ICL), Retrieval-Augmented
Generation (RAG), and Long-document QA (LongQA).

\subsection{Regularization Can Mitigate Length-Based Overfitting}\label{sec:mitigate-overfit}
Sec.~\ref{sec:long-train-hinder-lengen} showed that, as training progresses, the extrapolation of transformers trained with or without WD inevitably worsens. Here, we show that other regularization methods mitigate this effect.

Fig.~\ref{fig:eval-per-step-softmax} shows that dropout, specifically VPAD, mitigates length overfitting substantially.
Training for 80K steps does not bend the CE loss and RBA curves as much as it does for WD or no-WD. Moreover, the CE loss stays below both baselines at all training steps. On RULER, the dropout model already surpasses all baselines at 12K steps and, at 44K, outperforms the second-best model---no-WD---by $60\%$.
At the final training step (Fig.~\ref{fig:res-ppl-and-ruler-softmax}), all dropout variants improve the CE loss and allow the models to extrapolate much further on NIAH-1 and RULER, while the WD model collapses almost immediately at $2\times$ the ID length.
CPT of SmolLM3 (Table~\ref{tab:SmolLM3_niah_1}) gives a similar picture.
Neither the original model nor its SW variant extrapolates even on the simplest NIAH-1 task, and fine-tuning with WD or no-WD does not change this. In contrast, all dropout variants significantly mitigate the problem, allowing for non-zero performance even at $8\times$ the ID length.

Dropout is not the only regularization that improves extrapolation: we observe moderate gains (App.~\ref{app:additional-results}) for checkpoint averaging and attention dropout, which drops attention weights. This is reminiscent of classical \emph{overfitting}, but with respect to context length: a model fits its training distribution while its OOD performance degrades, and proper regularization mitigates it. Importantly, the effect is not specific to PE, as improvements hold for both NAPE and RoPE-SW models.

\subsection{Effect of Dropout Placement}\label{sec:exp-placement}
Sec.~\ref{sec:theory} predicts that the position of the dropout mask relative to normalization is decisive: Pre-Norm dropout attenuates the pre-activation mean (Proposition~\ref{prop:pn}). We hypothesize that the train-test mismatch is detrimental for extrapolation and that placing dropout just before the Linear layer (see Fig.~\ref{fig:dropout-placements}, Pre-Linear) is a better approach. Here, we test this hypothesis empirically.

\begin{figure}
    \centering
    \includegraphics[width=0.32\linewidth]{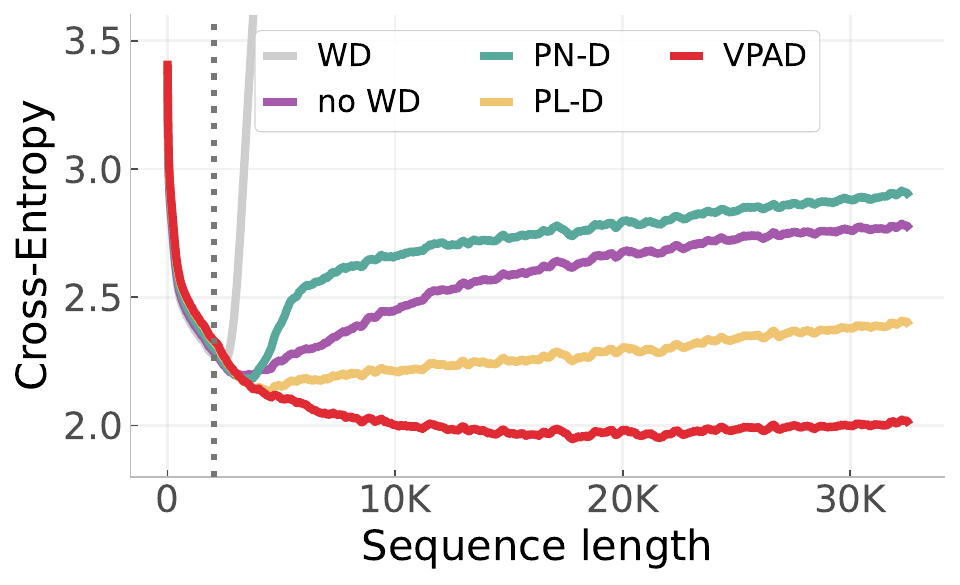}
    \includegraphics[width=0.32\linewidth]{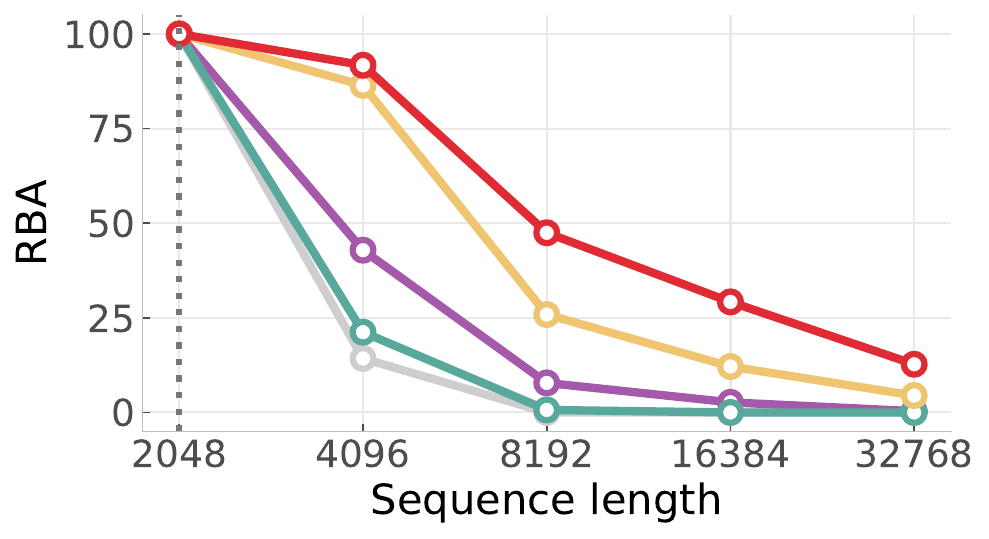}
    \includegraphics[width=0.32\linewidth]{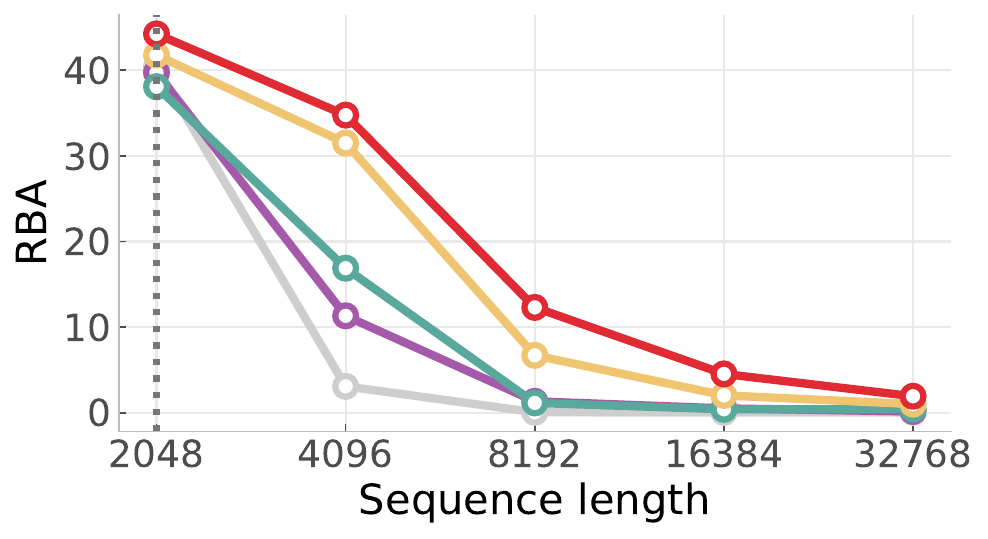}
    \caption{Results for transformers trained from scratch, evaluated on \textbf{Left:} ArXiv papers; \textbf{Middle:} NIAH-1; \textbf{Right:} RULER.
    The dotted lines represent the pre-trained context length.
    }
    \label{fig:res-ppl-and-ruler-softmax}
\end{figure}

Fig.~\ref{fig:res-ppl-and-ruler-softmax} shows the CE loss and the performance on NIAH-1 and on RULER for models trained from scratch. For CE and NIAH-1, Pre-Norm Dropout lags behind even the no-WD model, which justifies its disappearance from pre-training pipelines; in contrast, Pre-Linear Dropout is significantly better than WD, no-WD, and PN-D. The same ordering holds at the 3B scale under continual pre-training: in Table~\ref{tab:SmolLM3_niah_1} (full NIAH-1 in Table~\ref{tab:SmolLM3_niah_1-full}), PN-D extrapolates only partially to $2\times$ the ID length on NIAH-1 and collapses beyond it, whereas PL-D extrapolates perfectly, retaining 100.0 RBA up to $64\times$ (256K). The gap on overall RULER is equally consistent: PL-D more than doubles PN-D at $2\times$ extrapolation and remains operational at 4--8$\times$, where PN-D is near zero. In addition, dropout placement also affects sparse transformers (see Table~\ref{tab:asentmax-ruler} in App. \ref{app:asentmax-res}).

\subsection{Effect of Variance Preservation (VPAD)}\label{sec:exp-vpad}
With the placement fixed to Pre-Linear (Sec.~\ref{sec:pl}), we ask whether removing the remaining variance mismatch helps further by replacing standard inverted dropout with VPAD. When training from scratch, VPAD outperforms all other methods on all three evaluations of Fig.~\ref{fig:res-ppl-and-ruler-softmax}: its CE curve is the lowest and flattest at OOD lengths, and its RBA is the highest on both NIAH-1 and RULER at every context size---at 8K ($4\times$ extrapolation), it scores around 50 on NIAH-1 while the second-best model, PL-D, stays under 30. Although VPAD's CE is slightly higher in the 0--2K range, its ID performance does not suffer: at 2K, its RBA is the best among the five settings.

Under continual pre-training, VPAD nearly matches the perfect extrapolation of PL-D on NIAH-1, retaining 100.0 RBA up to $32\times$ and 99.2 at $64\times$ the ID length, and achieves the best RULER results at $2$--$8\times$ extrapolation. However, although OOD performance improves as the dropout probability increases, ID performance drops (Table~\ref{tab:SmolLM3_ruler_average}, CPT: VPAD); this also holds for PL-D and PN-D (Table~\ref{tab:SmolLM3_ruler_extended} in App.~\ref{app:smollm3-extended-ruler}), with VPAD degrading the least.
The effect is milder on standard short-context benchmarks (Table~\ref{tab:short-bench} in App.~\ref{app:results-on-short-benchs}): on-par or slightly better performance for $p=0.1$ and $0.2$, with minimal degradation for $p=0.3$.
In both cases, we presume the model unlearns some behaviors as it adapts to the new training dynamics, particularly for stronger dropout.
On HELMET at $2$--$8\times$ the ID length (Table~\ref{tab:SmolLM3_HELMET}), the SmolLM3-SW and WD baselines collapse to near-zero performance, while both dropout variants extrapolate: PL-D is better at 8K in ICL and LongQA, but VPAD is consistently better at $4$--$8\times$. Moreover, VPAD keeps improving on ICL and LongQA as the context grows, suggesting that it is more effective on much longer sequences. This is also evident from Table~\ref{tab:icl-helmet}, where its score increases until $32\times$ the pre-training length, whereas PL-D peaks at $4\times$.

\begin{table}[ht]
\centering
\vspace{-12pt}
\caption{SmolLM3 performance on NIAH-1 and on RULER.
Grey --- the ID length. Bold --- best. }
\label{tab:SmolLM3_niah_1}
\label{tab:SmolLM3_ruler_average}
\resizebox{0.78\linewidth}{!}{%
\begin{tabular}{lrrrcrcrrrr}
\toprule
& \multicolumn{5}{c}{NIAH-1} & & \multicolumn{4}{c}{RULER} \\
\cmidrule(lr){2-6} \cmidrule(lr){8-11}
Model & \cellcolor{idgray}4K  & 8K & 16K & 32-128K & 256K & & \cellcolor{idgray}4K  & 8K & 16K & 32K \\
\midrule
Orig. & \bf 100.0 & 0.0 & 0.0 &  --- 0.0 --- & 0.0 & & \bf 83.5 & 0.1 & 0.0 & 0.0 \\
SW & \bf 100.0 & 0.0 & 0.0 & --- 0.0 --- & 0.0 & & 83.4 & 0.5 & 0.0 & 0.0 \\
\dashmid
\bf CPT \\
SW, WD & \bf 100.0 & 0.0 & 0.0 & --- 0.0 --- & 0.0 & & \bf 83.5 & 1.1 & 0.0 & 0.0 \\
SW, no-WD & \bf 100.0 & 0.0 & 0.0 & --- 0.0 ---  & 0.0 & & \bf 83.5 & 1.1 & 0.0 & 0.0 \\
SW, PN-D $p=0.3$ &  \bf 100.0 & 82.2  & 6.2  & --- 0.0 --- & 0.0 & & 76.1 & 22.5 & 2.1  & 0.9 \\
SW, PL-D $p=0.3$ & \bf 100.0 & \bf 100.0 & \bf 100.0 & --- {\bf 100.0} --- & \bf 100.0 & & 74.1 & 50.6 & 40.2 & 25.5 \\
\dashmid
\multicolumn{3}{l}{\bf CPT: VPAD} \\
SW, VPAD $p=0.1$ & \bf 100.0 & \bf 100.0 & 57.2  & $30.4 \rightarrow 0.0$  & 0.0 & & 80.4 & 45.2 & 5.9 & 2.0 \\
SW, VPAD $p=0.2$ & \bf 100.0 & \bf 100.0 & \bf 100.0 & $96.2 \rightarrow 86.0$  & 49.4 & & 79.9 & \bf 55.0 & 38.8 & 19.4 \\
SW, VPAD $p=0.3$ & \bf 100.0 & \bf 100.0 & \bf 100.0 &  --- {\bf 100.0} --- & 99.2 & & 77.8 & 54.3 & \bf 44.1 & \bf 27.4 \\
\bottomrule
\end{tabular}%
}
\end{table}

\subsection{Dropout and Attention Scaling Are Complementary}\label{sec:exp-icl}

To improve extrapolation, we can apply length-dependent attention scaling, which has been shown to mitigate softmax-related issues \citep{chiang-cholak-2022-overcoming, nakanishi2025scalable, peng2023yarn}. In this section, we demonstrate that dropout and attention scaling are complementary. For evaluation, we use the ICL task from HELMET.
The prominent distinction of the ICL task is that the model has to learn from many examples (many-shot), and many LLMs benefit from it \citep{yen2025helmet}.
In this setting, we assume an extrapolation improvement when the model's score either increases or remains relatively stable across context lengths.
We show our results in Table~\ref{tab:icl-helmet}.

We first observe that Base is not capable of extrapolating $2\times$, which usually happens with RoPE models without RoPE extension. Even though YaRN gains the maximum score overall, it is not able to extrapolate far beyond $2\times$ the ID length.
In contrast, all SW-VPAD models improve even at 256K ($32\times$ extrapolation), and at 512K their degradation is negligible.  VPAD S$\times$ significantly surpasses other SW models at all sequence lengths and is superior at 512K overall. Moreover, the gap between VPAD and VPAD S$\times$ at 128--512K is significant---about 8 points. PL-D S$\times$ performance is less pronounced, but the same pattern holds: the scaling improves scores substantially.
This indicates that dropout and attention scaling are remedies to different extrapolation-related issues.

\begin{table}[t]
\begin{minipage}[t]{0.525\linewidth}
\centering
\caption{OOD performance of SmolLM3 and Mamba2 on ICL, RAG and LongQA from HELMET.
For SmolLM3, $p=0.3$ and, for Mamba2, $p=0.2$. Bold --- best.
}
\label{tab:SmolLM3_HELMET}
\resizebox{\linewidth}{!}{
\begin{tabular}{lrrrrrrrrr}
\toprule
& \multicolumn{3}{c}{ICL} & \multicolumn{3}{c}{RAG} & \multicolumn{3}{c}{LongQA}  \\
\cmidrule(lr){2-4} \cmidrule(lr){5-7} \cmidrule(lr){8-10}
Model & 8K & 16K & 32K & 8K & 16K & 32K & 8K & 16K & 32K\\
\midrule
\multicolumn{3}{l}{\bf Original Models} \\
SmolLM3-SW & 0.0 & 0.0 & 0.0 & 0.2 & 0.0 & 0.0 & 4.8 & 4.3 & 5.7 \\
Mamba2 &  8.4 & 4.0 & 4.0 & 5.4 & 2.0 & 1.6 & 7.3 & 4.5 & 3.2 \\
\dashmid
\multicolumn{3}{l}{\bf CPT: SmolLM3} \\
SW, WD & 0.0 & 0.0 & 0.0 & 0.3 & 0.0 & 0.0 & 6.3 & 6.7 & 6.7 \\
SW, PL-D & \bf 61.8 & 59.8 & 58.9 & 39.4 & 34.3 & 28.8 & \bf 17.7 & 15.2 & 16.8\\
SW, VPAD & 57.8 & \bf 61.5 & \bf 61.8 & \bf 40.2 & \bf 35.3 & \bf 31.0  & 14.9 & \bf 16.3 &  \bf 18.6 \\
\dashmid
\multicolumn{3}{l}{\bf CPT: Mamba2} \\
WD & 5.2 & 1.5 & 1.6 & 6.6 & 2.3 & 1.5 & \bf 12.6 &  9.2 & 7.5 \\
PL-D & \bf 20.5 & \bf 17.2  & \bf 9.8  & \bf 20.1 & \bf 17.1  & 11.2 & 10.8 & \bf 11.6  & \bf 12.1 \\
VPAD & 18.4 & 15.1  & 9.6 & 18.5 & 15.3  & \bf 12.1  & 9.3  & 8.4   & 9.8 \\
\bottomrule
\end{tabular}
}
\end{minipage}\hfill
\begin{minipage}[t]{0.45\linewidth}
\centering
\caption{Performance of SmolLM3-Base and SmolLM3-SW-8K (S$\times$) on ICL.
Bud. --- pre-training budget following stage 3.
Grey --- the ID length of the SW models.
Red --- the ID length of Base and YaRN.
$(.)^{*}$ --- YaRN factor 4 and $(.)^{**}$ --- YaRN factor 8. }
\label{tab:icl-helmet}
\resizebox{\linewidth}{!}{
\begin{tabular}{lrl;{0.4pt/1.6pt}rrrr}
\toprule
& \multicolumn{2}{c}{SmolLM3-Base} & \multicolumn{4}{c}{CPT: SmolLM3-SW-8K (Ours)} \\
\cmidrule(lr){2-3} \cmidrule(lr){4-7}
Len. & Base & YaRN & \multicolumn{1}{r}{PL-D} & PL-D S$\times$ & VPAD & VPAD S$\times$ \\
\midrule
\cellcolor{idgray} 8K & 54.5 & 48.1 & 60.5 & 61.9 & 58.3 & \bf 62.5 \\
16K & 65.6 & 61.9 & 61.8 & 64.4 & 61.7 & \bf 66.1 \\
32K & 73.2 & 70.2 & 63.6 & 66.3 & 65.9 & \bf 71.4 \\
\cellcolor{redrow} 64K & 78.2 & 76.9 & 62.3 & 67.4 & 66.1 & \bf 72.8 \\
128K & 4.6 & 83.3 & 62.1 & 68.8 & 66.1 & \bf 74.1 \\
256K & 0.0 & $77.2^{*}$ & 62.8 & 72.0 & 68.4 & \bf 76.6 \\
512K & 0.0 & $57.4^{**}$ & 63.2 & 69.2 & 68.0 & \bf 76.2 \\
\dashmid
Bud. & 100B & 100B & 1.5B & 1.5B & 1.5B & 1.5B \\
\bottomrule
\end{tabular}
}
\end{minipage}
\end{table}

\subsection{Mamba2}\label{sec:exp-mamba}
We next test whether our findings transfer to Mamba2---an architecture different from transformers.

\paragraph{Synthetic Copying}
Following \citet{buitrago2025understanding}, we conduct experiments on the copying task \citep{Jelassi2024RepeatAM}.
We report the mean Exact Match Accuracy at a sequence length of 150 (EMA-150) (EMA is used by \citet{Jelassi2024RepeatAM}) and Character-Level Accuracy at 300 (CLA-300) (as in \citet{buitrago2025understanding}) in Table~\ref{tab:res-mamba-copy}; standard deviations over five seeds are given in Table~\ref{tab:res-mamba-copy-std} (App.~\ref{app:additional-results}).
Table~\ref{tab:res-mamba-copy} shows that on average the results are consistent with our observation for transformers: WD worsens the length generalization ability and dropout improves it.

\paragraph{Training from scratch} Fig.~\ref{fig:mamba2-85m-ce-loss} shows the cross-entropy loss of the trained models. As with transformers, the performance of the WD model begins to degrade almost immediately beyond the pre-training length, indicated by the dotted line. The model trained without WD (no-WD) shows a better CE curve, but it still lags behind the dropout methods, which substantially flatten their curves, indicating improved extrapolation. In contrast to transformers, PL-D and VPAD perform similarly in terms of CE.
Furthermore, no-WD, PL-D, and VPAD improve performance on NIAH-1 (Table~\ref{tab:res-mamba-85m-niah-1}), with VPAD substantially outperforming all other baselines at $2$--$4\times$ extrapolation.

\paragraph{CPT} Table~\ref{tab:res-mamba-ruler} shows results on NIAH-1 and RULER (see Table~\ref{tab:res-mamba-ruler-full} for more results). As with transformers, extrapolation improves when Mamba2 is fine-tuned with dropout. On NIAH-1, PL-D and VPAD allow Mamba2 to perform perfectly at 16K and achieve non-zero performance at 32K, extending the extrapolation range $4\times$ over the original and non-dropout models. For RULER, all dropout variants improve length generalization substantially starting from the ID length. At $4\times$ the ID length, PL-D and VPAD double the score, and at $8$--$16\times$ they maintain non-zero performance while the original and non-dropout models fail completely.
PN-D mirrors the placement effect we observed for transformers (Sec.~\ref{sec:exp-placement}): it improves performance on RULER at $2\times$, but its gains at longer lengths are much less pronounced.
Further evidence for our regularization hypothesis comes from the HELMET results in Table~\ref{tab:SmolLM3_HELMET}, where we show $\geq4\times$ extrapolation performance.
The dropout variants substantially boost the results of ICL and RAG, giving a two- to ten-fold improvement over the original model.
Notably, VPAD interacts differently with Mamba2 than with transformers: VPAD is better on NIAH-1 and at the ID length of RULER, but it either lags behind or performs similarly to PL-D in extrapolation on both RULER and HELMET.

\begin{table}[ht!]
\centering
\begin{minipage}{0.54\linewidth}
    \centering
    \caption{
    Mamba2-45M and Mamba2-85M performance on Copy and NIAH-1, respectively.
    }
    \label{tab:res-mamba-copy}
    \label{tab:res-mamba-85m-niah-1}
    \resizebox{\linewidth}{!}{%
    \begin{tabular}{lrrcrrr}
        \toprule
        & \multicolumn{2}{c}{Copy} & & \multicolumn{3}{c}{NIAH-1} \\
        \cmidrule(lr){2-3} \cmidrule(lr){5-7}
        Model & EMA-150 & CLA-300 & & \cellcolor{idgray} 1K & 2K & 4K \\
        \midrule
        WD & 33.1 & 42.8 & & 72.3 & 30.5  & 1.1 \\
        no-WD & 42.7 & 45.7 & & \bf 91.2 & 50.6  & 1.1 \\
        PL-D & 61.5 & 50.9 & & 90.3 & 43.4 & 10.5 \\
        VPAD & \bf 68.9 & \bf 58.0 & & 88.9 & \bf 76.4  & \bf 30.2 \\
        \bottomrule
    \end{tabular}
    }
\end{minipage}
\hfill
\begin{minipage}{0.366\linewidth}
    \centering
    \includegraphics[width=0.99\linewidth]{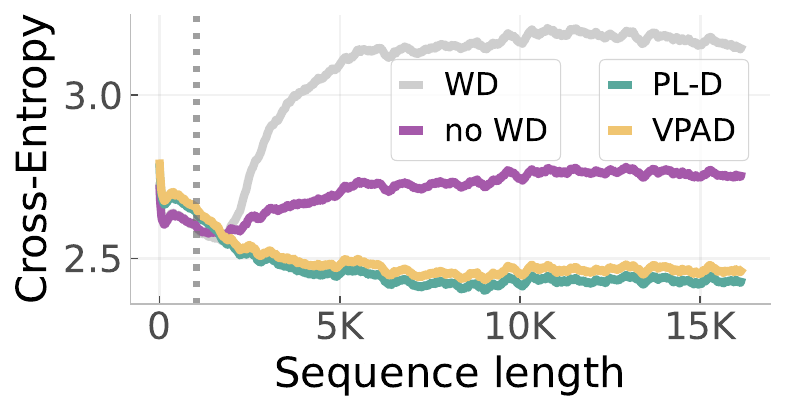}
    \captionof{figure}{CE loss of Mamba2-85M models on The Pile.
    }
    \label{fig:mamba2-85m-ce-loss}
\end{minipage}

\end{table}

\begin{table}[ht!]
\vspace{-12pt}
\centering
\caption{Mamba2-1.3B performance on NIAH-1 and on RULER. Bold --- best.
}
\label{tab:res-mamba-ruler}
\resizebox{0.8\linewidth}{!}{%
\begin{tabular}{lrrrrrcrrrrr}
\toprule
& \multicolumn{5}{c}{NIAH-1} & & \multicolumn{5}{c}{RULER} \\
\cmidrule(lr){2-6} \cmidrule(lr){8-12}
Model & \cellcolor{idgray}2K  & 4K & 8K & 16K & 32K & & \cellcolor{idgray}2K  & 4K & 8K & 16K & 32K \\
\midrule
Orig. & \bf 100.0 & \bf 100.0 & \bf 100.0 & 0.0 & 0.0 & & 38.3 & 14.3 & 8.5  & 0.4  & 0.0 \\
\dashmid
\bf CPT \\
WD & \bf 100.0 & \bf 100.0 & \bf 100.0 & 0.0 & 0.0 & & 42.0 & 16.2 & 9.6  & 0.7   & 0.0 \\
no-WD & \bf 100.0 & \bf 100.0 & \bf 100.0 & 0.0 & 0.0 & & 42.2 & 16.0 & 9.6  & 0.0   & 0.0 \\
PN-D, $p=0.2$ & \bf 100.0 & \bf 100.0 & \bf 100.0 & 11.8 & 0.0 & & 50.7 & 22.5 & 11.5 & 2.6 & 0.8 \\
PL-D, $p=0.2$ &  \bf 100.0 & \bf 100.0 & \bf 100.0 & \bf 100.0 & 22.8 & & 48.9 & \bf 28.8 & 18.0 & \bf 13.7  & 5.3 \\
VPAD, $p=0.2$ &  \bf 100.0 & \bf 100.0 & \bf 100.0 & \bf 100.0 & \bf 45.0 & & \bf 50.9 & 26.3 & \bf 18.5 & 13.3  & \bf 5.8 \\
\bottomrule
\end{tabular}%
}
\vspace{-9pt}
\end{table}

\section{Related Work}\label{sec:related-work}
Extrapolation failures in transformers are commonly attributed to PE \citep{press2022train, sun2023length, kazemnejad2023impact, chen2023extending, peng2023yarn, li2024functional} and to softmax attention \citep{chiang-cholak-2022-overcoming, velickovic2024softmax, barbero2024transformers, nakanishi2025scalable}.
\citet{buitrago2025understanding} show that linear RNNs overfit to length as training progresses.
We demonstrate that the same overfitting happens in transformers, is distinct from PE- and softmax- related issues, and can be mitigated by proper regularization.
\citet{du2025longshort} aligns output distributions across sequence lengths with an additional regularization term; we instead study standard regularization.
\citet{zhou2024transformers} observe that a substantial dropout rate ($0.2$) severely impairs extrapolation on integer addition, so their models use no dropout. We attribute this to the train-test mismatch analyzed here and reach the opposite conclusion: correctly placed dropout substantially improves length generalization.
\citet{wu-etal-2021-unidrop} and \citet{sankararaman2022bayesformertransformeruncertaintyestimation} explore placing dropout just before linear projections. Both target in-distribution performance and, importantly, do not analyze the mask's interaction with normalization.
In contrast, we show that the placement relative to the normalization layer determines the train-test mismatch of pre-activation statistics, which affects length generalization.
Inverted dropout removes the train-test gap of pre-activation means~\citep{srivastava2014dropout} but not of their variances~\citep{hendrycks2017adjusting, wager2013dropout}.
\citet{dopout-batch-norm-disharmony} identify a related dropout--batch-normalization mismatch in CNNs. We show the analogous mean mismatch in transformers combining dropout with layer normalization, and that extrapolation benefits from fixing the mismatch and from matching the variance. See more related work in App.~\ref{app:extended-related-work}.

\section{Conclusion}
We showed that transformers can progressively lose extrapolation ability during training, suggesting a form of length-based overfitting that proper regularization can substantially mitigate.
In that regard, we showed that standard Pre-Norm dropout introduces a train-test mismatch in pre-activation statistics, Pre-Linear dropout removes the mean mismatch, and VPAD additionally preserves the marginal variance. These changes substantially improve extrapolation, including perfect NIAH recall up to $64\times$ the training context in our continual-pretraining setting.
Our results therefore suggest that length generalization is affected not only by the architecture and positional encoding but also by the choice of regularization. The right choice of dropout can mitigate length-based overfitting without introducing new train-test distribution shifts.

\section*{Acknowledgments}

We thank the SARDINE Lab members for reviewing this paper and providing helpful feedback.
This work was supported by the Portuguese Recovery and Resilience Plan through project C645008882-00000055 (Center for ResponsibleAI), by the project DECOLLAGE (ERC-2022-CoG 101088763), and by FCT/MECI through national funds and when applicable co-funded EU funds under DOI:10.54499/UID/50008/2025: Instituto de Telecomunicações.
We thank EuroHPC for the HPC resources used to support this work through grants EHPC-AI-2024A05-044 and E-EXT-2025E01-100.

\clearpage
\newpage

\bibliographystyle{iclr2027_conference}
\bibliography{refs}

\appendix

\clearpage
\newpage

\section{Extended Related Work}\label{app:extended-related-work}
\paragraph{Length generalization in transformers.}
\citet{varis-bojar-2021-sequence} identify sequence length as a domain to which transformers overfit, and \citet{anil2022exploring} show that even large models fail to extrapolate. Extrapolation failures are commonly attributed to the positional encoding: relative and bias-based encodings improve extrapolation---relative position representations \citep{shaw2018self, raffel2020exploring}, ALiBi \citep{press2022train}, KERPLE \citep{chi2022kerple}, Sandwich \citep{chi2023dissecting}, and xPos \citep{sun2023length}---while later encodings learn, adapt, or redesign the positional map \citep{li2024functional, zheng2024dape, golovneva2024contextual, he2024two, hua2025fourier, wang2024resonance, yang2025path}. \citet{kazemnejad2023impact} show that decoder-only models can extrapolate better with no positional encoding at all, and \citet{ruoss2023randomized} randomize training positions to cover longer test lengths. The second commonly named cause is softmax attention, which disperses as the input grows: \citet{chiang-cholak-2022-overcoming} show that this dilution breaks extrapolation and counteract it with a log-length scaling of the logits, \citet{velickovic2024softmax} argue that dispersion is fundamental to softmax, and \citet{nakanishi2025scalable} propose Scalable-Softmax as a replacement. \citet{barbero2024transformers} analyze over-squashing and representational collapse on long sequences. Our experiments with models trained from scratch use NAPE \citep{vasylenko2026longcontext}, a NoPE + ALiBi strategy, and the CPT ones use RoPE inside sliding-window attention layers. Moreover, we employ input- and length-dependent attention scaling.
We show that regularization provides further gains on top of a proper PE strategy (Sec.~\ref{sec:mitigate-overfit}) and attention scaling (Sec.~\ref{sec:exp-icl}), indicating that PE and attention are not the whole story.

\paragraph{Extending the context window of pre-trained models.}
Many works adapt pre-trained models to longer contexts: position interpolation
\citep{chen2023extending}, frequency-aware RoPE rescaling
\citep{peng2023yarn}, scaling laws and theoretical bounds for RoPE extrapolation
\citep{liu2024scaling, xu2024base}, LongRoPE \citep{ding2024longrope},
CLEX \citep{chen2024clex}, PoSE \citep{zhu2024pose}, and long-context
training recipes \citep{xiong2024effective, fu2024data,
gao-etal-2025-train, chen2024longlora}.
Training-free methods modify positional information or attention computation
at inference \citep{jin2024llm, an2024training, ma2024mesa, an2025effective}
or use sliding-window attention with retained initial tokens
\citep{xiao2024efficient, han2024lminfinite}, following earlier
sparse/local-attention architectures
\citep{child2019generating, beltagy2020longformer, jiang2023mistral}.
Our CPT experiments adopt sliding-window attention in the RoPE layers and rely
on regularization rather than positional remapping. Importantly, we do not fine-tune on longer sequences.

\paragraph{Regularization for length generalization.}
\citet{du2025longshort} study length generalization from the perspective of model output distributions. They introduce Long-Short Misalignment, which measures the discrepancy between output distributions across sequences of different lengths, and improve length generalization by explicitly regularizing this discrepancy during training.
\citet{zhou2024transformers} investigate length generalization on integer addition and show that it is highly sensitive to training conditions. In particular, they observe that a dropout rate of $0.2$ severely impairs length generalization, while $0.1$ provides no significant improvement, and consequently train their models without dropout.
In contrast, we study how standard regularization choices themselves affect length-based overfitting in language models. We show that dropout can substantially improve length generalization when its placement is reconsidered, and identify the train-test mismatch induced by conventional dropout placement as an important factor. Unlike \citet{du2025longshort}, our approach does not explicitly enforce consistency across different sequence lengths and does not require two forward passes.

\paragraph{Length generalization in recurrent and state-space models.}
Fixed-state models face their own extrapolation limits. \citet{buitrago2025understanding} attribute failures of linear RNNs to overfitting to the recurrent states explored during training, \citet{ye2025longmamba} to global channels failing to adaptively extend
their receptive fields beyond the training length, and \citet{Jelassi2024RepeatAM} show that copying is harder without attention. Empirically, such models lag transformers on tasks requiring in-context recall \citep{waleffe2024empirical, arora2024zoology}. Proposed remedies include token decimation \citep{ben-kish2025decimamba}, receptive-field enlargement \citep{ye2025longmamba}, spectrum scaling of the state-transition dynamics \citep{lu2025mamba}, chunked inference that prevents state overflow \citep{ben-kish2025overflow}, and tool use \citep{malach2026infinity}. Our Mamba2 experiments show that proper regularization can also improve extrapolation.

\paragraph{Relation to Alpha-Dropout.} The affine correction underlying VPAD coincides with that of Alpha-Dropout~\citep{klambauer2017selfnormalizing}, which likewise restores both the mean and the variance of a masked activation through a deterministic scale and shift. On one hand, VPAD is the special case in which dropped units are set to zero rather than to the SELU saturation value $\alpha'$ (i.e.\ $\alpha' = 0$). On the other hand, unlike Alpha-Dropout, VPAD makes no zero-mean and unit-variance assumption on the activation. VPAD is the general, non-centered case that arises under RMSNorm.

\section{Proofs}\label{app:proofs-rms}

Throughout this appendix we use the notation of Sec.~\ref{sec:preact-setup}: $\vv{x} \in \R^d$ has i.i.d.\ entries with mean $\mu_x$ and variance $\sigma_x^2$; the masks $b_i \sim \mathrm{Bernoulli}(1-p)$ are independent of $\vv{x}$, where $p \in [0,1)$ is the drop probability; and the normalized coordinate $\hat{x}_i = x_i/(\RMS(\vv{x}) + \epsilon)$. Expectations are taken over all sources of randomness---the input activations and the dropout masks, which are independent of each other.

\subsection{Moments of the normalized coordinate and inference baseline}\label{app:proofs-baseline}

We work in the large-$d$ regime, where the relevant sample averages are concentrated around their expectations.
Write $y^{\mathrm{test}} = \sum_i w_i\,\hat{x}_i$.
By the strong law of large numbers, $\RMS(\vv{x})^2 = \tfrac{1}{d}\sum_j x_j^2 \to \mu_x^2 + \sigma_x^2$ almost surely. Since $r \mapsto 1/(\sqrt{r}+\epsilon)$ is continuous on $[0,\infty)$, the continuous mapping theorem gives $\hat{x}_i = x_i/(\RMS(\vv{x}) + \epsilon) \to x_i/(\sqrt{\mu_x^2+\sigma_x^2} + \epsilon)$ almost surely. Next, since $|\hat{x}_i| \le |x_i|/\epsilon$ and $\hat{x}_i^2 \le x_i^2/\epsilon^2$ are integrable envelopes that do not depend on $d$ (as $\RMS(\vv{x}) + \epsilon \ge \epsilon$), dominated convergence ($\lim_{d\to\infty} \E[\hat{x}_i] = \E[\lim_{d\to\infty} \hat{x}_i]$, and likewise for $\hat{x}_i^2$) gives
\begin{equation}\label{eq:x-hat-limits}
\mu_{\hat{x}} \to \frac{\mu_x}{\sqrt{\mu_x^2 + \sigma_x^2} + \epsilon},
\qquad
\E[\hat{x}_i^2] \to \frac{\mu_x^2 + \sigma_x^2}{(\sqrt{\mu_x^2 + \sigma_x^2}+\epsilon)^2},
\qquad
\sigma_{\hat{x}}^2 = \E[\hat{x}_i^2] - \mu_{\hat{x}}^2 \to \frac{\sigma_x^2}{(\sqrt{\mu_x^2 + \sigma_x^2} + \epsilon)^2}.
\end{equation}

This proves the per-coordinate limits, and linearity of expectation gives the mean claim of Eq.~\eqref{eq:baseline}: $\E[y^{\mathrm{test}}] = \sum_i w_i\,\E[\hat{x}_i] = \mu_{\hat{x}} \sum_i w_i$.
By exchangeability, every coordinate has the same variance $\sigma_{\hat{x}}^2$, hence:
\[
\Var[y^{\mathrm{test}}]
\;=\; \sum_i w_i^2\,\Var[\hat{x}_i]
\;+\; \sum_{i \neq j} w_i w_j\,\Cov[\hat{x}_i, \hat{x}_j]
\;=\; \sigma_{\hat{x}}^2\,\|\vv{w}\|^2 \;+\; C(\hat{\vv{x}}, \vv{w}),
\]
where $C(\hat{\vv{x}}, \vv{w})=\sum_{i \neq j} w_i w_j\,\Cov[\hat{x}_i, \hat{x}_j]$ and $\Cov[\hat{x}_i, \hat{x}_j]$
is the cross-coordinate covariance induced by the shared normalization denominator.
Note that Eq. \ref{eq:baseline} also holds for non-i.i.d. $\vv{x}$
and non-exchangeable $\hat{\vv{x}}$. It only requires that the $\hat{x}_i$ share the same mean and variance. Thus, the observations in Sec.~\ref{sec:pl} and \ref{sec:variance-preserve-drop} also hold in that case.

\subsection{Proof of Proposition~\ref{prop:pn} (Pre-Norm dropout)}\label{app:proof-pn}

Recall from Sec.~\ref{sec:pn} that under Pre-Norm dropout the normalization input decomposes as $\vv{x} = \vv{r} + \vv{u}$, with $\vv{u}$ (mean $\mu_u$, variance $\sigma_u^2$) the sublayer output and $\vv{r}$ the residual; the masked input and normalized coordinate are $\tilde{x}_i = r_i + \tfrac{b_i}{1-p}u_i$ and $z_i^{\mathrm{PN}} = \tilde{x}_i/(\RMS(\tilde{\vv{x}}) + \epsilon)$, Eq.~\eqref{eq:pn-input}. We take the pairs $(r_i, u_i)$ i.i.d.\ across coordinates and independent of the masks $\{b_j\}$, with no independence assumed between $r_i$ and $u_i$; throughout, $\E[b_i] = \E[b_i^2] = 1-p$. Write $\kappa = \E[u_i^2]/\E[x_i^2] = \E[u_i^2]/(\mu_x^2 + \sigma_x^2)$ as in Sec.~\ref{sec:pn}.

\paragraph{Moments of the masked input.} Inverted dropout preserves the mean,
\[
\E[\tilde{x}_i] = \E[r_i] + \frac{\E[b_i]}{1-p}\E[u_i] = \E[r_i] + \E[u_i] = \mu_x ,
\]
and the masked second moment is
\begin{align*}
\E[\tilde{x}_i^2]
&= \E[r_i^2] + 2\,\frac{\E[b_i]}{1-p}\,\E[r_i u_i] + \frac{\E[b_i^2]}{(1-p)^2}\,\E[u_i^2] \\
&= \underbrace{\E[r_i^2] + 2\,\E[r_i u_i] + \E[u_i^2]}_{=\,\E[(r_i+u_i)^2]\,=\,\mu_x^2+\sigma_x^2} + \Bigl(\tfrac{1}{1-p}-1\Bigr)\E[u_i^2]
\;=\; (\mu_x^2+\sigma_x^2)\Bigl(1+\tfrac{p}{1-p}\,\kappa\Bigr) ,
\end{align*}
using $\tfrac{1}{1-p}-1 = \tfrac{p}{1-p}$ and $\E[u_i^2] = \kappa\,(\mu_x^2+\sigma_x^2)$. The cross term $2\,\E[r_i u_i]$ is identical at training and inference---the inverted-dropout factor restores the mask expectation to one---so the correlation between $r_i$ and $u_i$ cancels and only the diagonal term $\E[u_i^2]$ is inflated. Since the mean is preserved, $\Var[\tilde{x}_i] = \E[\tilde{x}_i^2] - \mu_x^2 = \sigma_x^2 + \tfrac{p}{1-p}\,\kappa\,(\mu_x^2+\sigma_x^2)$.

\paragraph{Normalized coordinate.} By the strong law of large numbers, $\RMS(\tilde{\vv{x}})^2 = \tfrac{1}{d}\sum_j \tilde{x}_j^2 \to (\mu_x^2+\sigma_x^2)\bigl(1+\tfrac{p}{1-p}\,\kappa\bigr)$ almost surely as $d \rightarrow \infty$, while at inference $\RMS(\vv{x})^2 \to \mu_x^2+\sigma_x^2$. Hence $z_i^{\mathrm{PN}} = \tilde{x}_i/(\RMS(\tilde{\vv{x}}) + \epsilon) \to \tilde{x}_i/\bigl(\sqrt{(\mu_x^2+\sigma_x^2)(1+\tfrac{p}{1-p}\,\kappa)}+\epsilon\bigr)$ almost surely, and since $|z_i^{\mathrm{PN}}| \le |\tilde{x}_i|/\epsilon$ is an integrable dominating envelope that does not depend on $d$, dominated convergence yields
\[
\E[z_i^{\mathrm{PN}}] \to \frac{\mu_x}{\sqrt{(\mu_x^2+\sigma_x^2)\bigl(1+\tfrac{p}{1-p}\,\kappa\bigr)}+\epsilon},
\qquad\text{and, at inference,}\qquad
\E[\hat{x}_i] \to \frac{\mu_x}{\sqrt{\mu_x^2+\sigma_x^2}+\epsilon}.
\]

\paragraph{Aggregation through the linear layer.} The mean claim needs only linearity: $\E[y^{\mathrm{PN}}] = \E[z_i^{\mathrm{PN}}]\sum_i w_i$ and $\E[y^{\mathrm{test}}] = \E[\hat{x}_i]\sum_i w_i$, so for $\mu_x \neq 0$ and any weights with $\sum_i w_i \neq 0$ the ratio is free of $\vv{w}$ at every finite $d$ and
\[
\frac{\E[y^{\mathrm{PN}}]}{\E[y^{\mathrm{test}}]} \;=\; \frac{\E[z_i^{\mathrm{PN}}]}{\E[\hat{x}_i]} \;\to\; \frac{\sqrt{\mu_x^2+\sigma_x^2}+\epsilon}{\sqrt{(\mu_x^2+\sigma_x^2)\bigl(1+\tfrac{p}{1-p}\,\kappa\bigr)}+\epsilon} \;=\; c_\epsilon
\;=\; \Bigl(1+\tfrac{p}{1-p}\kappa\Bigr)^{-1/2} + O(\epsilon)
\]
(when $\mu_x = 0$, both $\E[z_i^{\mathrm{PN}}]$ and $\E[\hat{x}_i]$ tend to $0$ and there is no mean mismatch to speak of). \qed

\subsection{Pre-Linear dropout}\label{app:proof-pl}

The mismatch of Pre-Linear dropout, Eq.~\eqref{eq:pl-moments}, is a classical property of inverted dropout, so we state and prove it here rather than in the main text.

\begin{proposition}[Pre-Linear dropout]\label{prop:pl}
The pre-activation moments satisfy
\begin{equation}
\E[y^{\mathrm{PL}}] \;=\; \E[y^{\mathrm{test}}],
\qquad
\Var[y^{\mathrm{PL}}] \;=\; \Var[y^{\mathrm{test}}] \;+\; \frac{p}{1-p}\,\E[\hat{x}_i^2]\,\|\vv{w}\|^2.
\end{equation}
\end{proposition}

Proof. By independence of $b_i$ and $\hat{x}_i$, and using $\E[b_i] = \E[b_i^2] = 1-p$,
\[
\E[z_i^{\mathrm{PL}}] = \frac{1}{1-p}\,\E[b_i]\,\E[\hat{x}_i] = \mu_{\hat{x}}, \qquad
\E[(z_i^{\mathrm{PL}})^2] = \frac{1}{(1-p)^2}\,\E[b_i^2]\,\E[\hat{x}_i^2] = \frac{\E[\hat{x}_i^2]}{1-p}.
\]
Hence, since $\E[\hat{x}_i^2] - \mu_{\hat{x}}^2 = \sigma_{\hat{x}}^2$,
\[
\Var[z_i^{\mathrm{PL}}] \;=\; \frac{\E[\hat{x}_i^2]}{1-p} - \mu_{\hat{x}}^2 \;=\; \sigma_{\hat{x}}^2 \;+\; \frac{p}{1-p}\,\E[\hat{x}_i^2].
\]
Moreover, the cross-coordinate covariances are preserved exactly: for $i \neq j$, since $(b_i, b_j)$ are independent of $(\hat{x}_i, \hat{x}_j)$ and of each other,
\[
\Cov[z_i^{\mathrm{PL}}, z_j^{\mathrm{PL}}] \;=\; \frac{\E[b_i]\,\E[b_j]}{(1-p)^2}\,\E[\hat{x}_i\hat{x}_j] - \E[\hat{x}_i]\,\E[\hat{x}_j] \;=\; \Cov[\hat{x}_i, \hat{x}_j]:
\]
the $1/(1-p)$ rescaling exactly undoes the mask's attenuation of the covariances, so the covariance terms of $\Var[y^{\mathrm{PL}}]$ and $\Var[y^{\mathrm{test}}]$ are both equal to $C(\vv{\hat{x}}, \vv{w})$ and cancel in the difference. Aggregation through the linear layer therefore gives
\[
\E[y^{\mathrm{PL}}] = \E[y^{\mathrm{test}}], \qquad
\Var[y^{\mathrm{PL}}] = \Var[y^{\mathrm{test}}] + \frac{p}{1-p}\,\E[\hat{x}_i^2]\,\|\vv{w}\|^2. \qed
\]

\subsection{Variance-Preserving Affine Dropout}\label{app:proof-darms}

The scale and shift $\alpha,\beta$ of~\eqref{eq:vpad-coef} are the affine correction of Alpha-Dropout~\citep{klambauer2017selfnormalizing} with dropped value $\alpha'=0$, generalized to a non-centered activation; we state the moment-matching property and prove it here rather than in the main text.

\begin{proposition}[Variance-Preserving Affine Dropout]\label{prop:darms}
With the scale $\alpha$ and shift $\beta$ of~\eqref{eq:vpad-coef}, the corrected activation~\eqref{eq:vpad-coef} preserves both moments of the normalized activation, $\E[z_i^{\mathrm{VPAD}}] = \mu_{\hat{x}}$ and $\Var[z_i^{\mathrm{VPAD}}] = \sigma_{\hat{x}}^2$; consequently
\begin{equation}
\E[y^{\mathrm{VPAD}}] \;=\; \E[y^{\mathrm{test}}],
\qquad
\Var[y^{\mathrm{VPAD}}] \;=\; \Var[y^{\mathrm{test}}] \;+\; \bigl(\alpha^2(1-p)^2 - 1\bigr)\,C(\vv{\hat{x}}, \vv{w}).
\end{equation}
\end{proposition}

Proof. Throughout, $\hat{x}_i$ is independent of $b_i$, with $\E[b_i] = \E[b_i^2] = 1-p$, and $\E[\hat{x}_i] = \mu_{\hat{x}}$, $\Var[\hat{x}_i] = \sigma_{\hat{x}}^2$, hence $\E[\hat{x}_i^2] = \sigma_{\hat{x}}^2 + \mu_{\hat{x}}^2$.

\paragraph{Mean.} Since $\beta$ is deterministic and $\hat{x}_i \perp b_i$,
\[
\E[z_i^{\mathrm{VPAD}}] \;=\; \alpha\,\E[b_i]\,\E[\hat{x}_i] + \beta \;=\; \alpha\,(1-p)\,\mu_{\hat{x}} + \beta.
\]
Substituting $\beta = \mu_{\hat{x}}\,\bigl(1 - \alpha\,(1-p)\bigr)$ yields $\E[z_i^{\mathrm{VPAD}}] = \mu_{\hat{x}}$.

\paragraph{Variance.} Since $\beta$ is constant,
\begin{align*}
\Var[\hat{x}_i\,b_i]
&= \E[b_i^2]\,\E[\hat{x}_i^2] - \bigl(\E[b_i]\,\E[\hat{x}_i]\bigr)^2 \\
&= (1-p)(\sigma_{\hat{x}}^2 + \mu_{\hat{x}}^2) - (1-p)^2 \mu_{\hat{x}}^2 \;=\; (1-p)\,(\sigma_{\hat{x}}^2 + p\,\mu_{\hat{x}}^2),
\end{align*}
using $1 - (1-p) = p$. Therefore $\Var[z_i^{\mathrm{VPAD}}] = \alpha^2\,(1-p)\,(\sigma_{\hat{x}}^2 + p\,\mu_{\hat{x}}^2)$; setting this equal to $\sigma_{\hat{x}}^2$ gives $\alpha^2 = \sigma_{\hat{x}}^2/\bigl((1-p)(\sigma_{\hat{x}}^2 + p\,\mu_{\hat{x}}^2)\bigr)$, well-defined and positive since $\sigma_{\hat{x}}^2 + p\,\mu_{\hat{x}}^2 > 0$ for any non-degenerate activation.

\paragraph{Aggregation, with the covariance term.} For the pre-activation mean, linearity gives $\E[y^{\mathrm{VPAD}}] = \mu_{\hat{x}}\sum_i w_i = \E[y^{\mathrm{test}}]$. For the variance, the covariance term must be tracked: for $i \neq j$, since $(b_i, b_j)$ are independent of $(\hat{x}_i, \hat{x}_j)$ and of each other,
\[
\Cov[z_i^{\mathrm{VPAD}}, z_j^{\mathrm{VPAD}}] \;=\; \alpha^2\,\E[b_i]\,\E[b_j]\,\bigl(\E[\hat{x}_i\hat{x}_j] - \E[\hat{x}_i]\,\E[\hat{x}_j]\bigr) \;=\; \alpha^2(1-p)^2\,\Cov[\hat{x}_i, \hat{x}_j]:
\]
every pairwise covariance is attenuated by the fixed factor $\alpha^2(1-p)^2 = \tfrac{\sigma_{\hat{x}}^2(1-p)}{\sigma_{\hat{x}}^2+p\,\mu_{\hat{x}}^2} \in (0,1)$, while at inference the covariances enter at full strength. \qed

\section{Experimental Details}\label{appendix:experiment-details}

\subsection{Experimental Setup}\label{sec:experimental-setup-full}

\begin{table}[]
    \centering
    \caption{Hyperparameters of the trained models. Ls --- number of layers, Dim --- model's dimension, FFN-D --- hidden dimension of FFN sublayers, H --- number of heads, Len --- context length, TT --- training tokens, BT --- tokens per batch, LR --- learning rate, LRS --- LR scheduler, Ann-R --- annealing ratio, min-LR --- annealing min LR. LR schedulers: Stable with Warm-Up and Annealing (SWUA), Stable with Warm-Up (SWU).}
    \resizebox{\linewidth}{!}{%
    \begin{tabular}{lcccccccccccccc}
    \toprule
Model & $p$ & Ls & Dim & FFN-D & H & Len & Data & TT &  BT & LR & LRS & Warm-up & Ann-R & min-LR \\
    \midrule
    \bf Trained from scratch & \\
Transformers, NAPE & 0.1 & 24 & 768 & 3072 & 12 & 2048 & DCLM-Edu & 40B & 512K & $6\cdot10^{-4}$ & SWUA & $2.5\%$ & $20\%$ & $6\cdot10^{-5}$ \\
Mamba2-85M, State=128 & 0.1 & 12 & 768 & - & 24 & 1024 & The Pile & 25B & 512K & $1\cdot10^{-3}$ & Cosine & $10\%$ & - & $1\cdot10^{-4}$ \\
Mamba2-45M, State=128 & 0.1 & 12 & 768 & - & 24 & 50-100 & Copy & 0.3B & 19K & $1\cdot10^{-3}$ & Cosine & $5\%$ & - & 0 \\
\dashmid
\bf CPT \\
SmolLM3, SW & 0.1-0.3 & 36 & 2048 & 11008 & 16 & 4096 & DCLM-Edu & 1B & 1M & $2\cdot10^{-5}$ & SWU & $10\%$ & - & $2\cdot10^{-5}$ \\
SmolLM3-8K, SW & 0.3 & 36 & 2048 & 11008 & 16 & 8192 & DCLM-Edu & 1.5B & 1M & $2\cdot10^{-5}$ & SWU & $10\%$ & - & $2\cdot10^{-5}$ \\
Mamba2-1.3B, State=128 & 0.1-0.3 & 48 & 2048 & - & 64 & 2048 & DCLM-Edu & 0.5B & 256K & $1\cdot10^{-4}$ & Cosine & $10\%$ & - & $1\cdot10^{-5}$ \\
\bottomrule
    \end{tabular}
}
    \label{tab:experiment-details}
\end{table}

\paragraph{Dropout implementation.} As discussed in Sec.~\ref{sec:theory}, Pre-Norm dropout is applied before RMSNorm and thus introduces one source of randomness per sublayer. Pre-Linear and VPAD, however, can have either one or several sources of randomness: a single mask after RMSNorm output or an independent mask per linear projection. For a fair comparison, we apply dropout once after RMSNorm, so that Pre-Linear and VPAD also have one source of randomness per sublayer. Future work could explore applying the mask per linear layer to introduce more randomness.

\paragraph{Training from scratch.} For experiments with transformers, we use a Llama-3-like model but replace RoPE with NAPE \citep{vasylenko2026longcontext}, which has shown its effectiveness at extrapolation without any post-training intervention; we use no attention scaling or any other intervention related to positional encoding. We use the DCLM-Edu \citep{allal2025smollm2smolgoesbig} dataset for training. We train three 400M-parameter models with different initializations for each of five settings: weight decay (WD), no weight decay (no-WD), Pre-Norm Dropout (PN-D), Pre-Linear Dropout (PL-D), and VPAD. Fig.~\ref{fig:eval-per-step-softmax} reports results for the best-performing model, and Fig.~\ref{fig:res-ppl-and-ruler-softmax} reports the average over 3 seeds.
The best model is selected based on the average RULER performance at 4--8K. We show the best-performing model to demonstrate that even the best WD model overfits badly when trained longer. The overall picture can be concluded from Fig.~\ref{fig:res-ppl-and-ruler-softmax}, which shows the average performance: for instance, no-WD overfits more on average.

\paragraph{Continual Pre-training.}
In order to test dropout at scale, we resort to continual pre-training of larger models. For transformers, we use the SmolLM3-3B checkpoint (stage 3, step 4720000), pre-trained with a context length of 4096 tokens. We convert its RoPE layers to sliding-window (SW) attention  with a window size of 4096 while leaving the NoPE layers untouched (i.e. they use global attention). The model is then trained with the same 4096 context size on about 1B tokens.
We evaluate the models on NIAH-1, RULER \citep{hsieh2024ruler} and a subset of HELMET \citep{yen2025helmet} tasks: In-Context Learning (ICL), Retrieval-Augmented
Generation (RAG), and Long-document QA (LongQA).

\paragraph{CPT: complementarity of dropout and attention scaling.}
We train a SmolLM3 model with an 8K context size and a 4K sliding window on 1.5B tokens.
Training on an 8K context allows the model to learn the sliding-window behavior better. It also allows us to report ID performance, since the HELMET evaluation starts at a sequence length of 8K.
The probability of dropout is set to $p=0.3$. We call this model SmolLM3-SW-8K and the scaled variant --- SmolLM3-SW-8K S$\times$. For the scaled model, we use an input-and-length-dependent scaling function introduced in \citet{vasylenko2026longcontext}. We compare the performance of our models with the original SmolLM3-Base \citep{bakouch2025smollm3} and SmolLM3-Base with the YaRN RoPE extension. The released SmolLM3-Base checkpoint was trained on around 100B additional tokens after the stage 3 checkpoint (the checkpoint that we use in all our experiments with SmolLM3).
We evaluate these models on the ICL tasks from HELMET and generate additional 256K-length and 512K-length sets (only possible for ICL tasks) with 100 samples per task (so 500 in total per set).

For all experiments, we use AdamW with $\beta_1=0.9$, $\beta_2=0.95$, and gradient clipping of 1.0. All  experiments labeled "WD" use weight decay 0.1 and 0 otherwise.
All SW models are trained with a sliding window of 4096. We train models from scratch with TorchTitan \citep{liang2025torchtitan} and train the CPT models with the Hugging Face Transformers library \citep{wolf-etal-2020-transformers}.
Additional hyperparameters are given in Table~\ref{tab:experiment-details}.

\subsection{Mamba2}

\paragraph{Dropout placement} Mamba2 blocks have two linear projections: the input projection, $\mM{W}_{\mathrm{in}}$, and the output projection. To match the transformer experiments as closely as possible, we work with $\mM{W}_{\mathrm{in}}$: like $W_{\{Q,K,V,G,U\}}$ in the attention and FFN sublayers, it is applied directly to the normalized input. Since the order of RMSNorm and the input projection is the same as in transformers, we can test all dropout variants. The implementation is also identical: dropout is patched into the RMSNorm module rather than attached to $\mM{W}_{\mathrm{in}}$.

\paragraph{Synthetic Copying}
Following \citet{buitrago2025understanding}, we test our findings on the copying task \citep{Jelassi2024RepeatAM}. The source sequence length during training ranges from 50 to 100 tokens, and the number of training samples is 5M. We train five models with different random seeds for each setting. We report the mean Exact Match Accuracy at a sequence length of 150 (EMA-150) and Character-Level Accuracy at 300 (CLA-300) (as in \citet{buitrago2025understanding}) in Table~\ref{tab:res-mamba-copy}; standard deviations over five seeds are given in Table~\ref{tab:res-mamba-copy-std} (App.~\ref{app:additional-results}). We chose sequences of length 150 because the Mamba2 models barely generalize to this length, while at length 200, all models achieve zero performance. This is consistent with the results reported by \citet{Jelassi2024RepeatAM}.

\paragraph{Training from scratch} In the case of Mamba2, we follow \citet{buitrago2025understanding} and train Mamba2 models with 85M parameters, also with three random initializations per setting. The training budget of the Mamba2 models is 25B tokens, as the degradation of length generalization is already noticeable at this stage.

\paragraph{CPT} We apply CPT to a 1.3B-parameter Mamba2 model, training for 0.5B tokens.

\subsection{Evaluation}

We evaluate models' performance via CE loss and Recall-Based Accuracy (RBA) on the Needle-in-a-Haystack task (NIAH-1 of RULER, also known as Passkey Retrieval) and on overall RULER.
To evaluate transformers' CE loss, we use 2000 long documents from ArXiv papers---a subset of The Common Pile dataset \citep{kandpal2025common}---and, for Mamba2, we follow \citet{buitrago2025understanding} by using The Pile \citep{pile}.
We use the LM Evaluation Harness \citep{eval-harness} for the short-context benchmarks and RULER. For HELMET \citep{yen2025helmet}, we use its official implementation.

\section{Additional empirical results}\label{app:additional-results}

\subsection{Results on Short Benchmarks}\label{app:results-on-short-benchs}
We report performance on standard short-context benchmarks in Table~\ref{tab:short-bench}.

\begin{table}[]
    \centering
    \small
     \caption{SmolLM3-3B results on short benchmarks. HS - HellaSwag, WG - Winogrande, LMBD - Lambada, OPQA - OpenBookQA}
    \label{tab:short-bench}
\begin{tabular}{lcrrrrrrrrr}
\toprule
Model & $p$ & ARC-E & ARC-C & HS & PIQA & WG & OPQA & BoolQ & LMBD & Avg. \\
\midrule
Original & - & 77.8 & 53.2 & 75.3 & 79.0 & 69.6 & 44.2 & 80.4 & 70.8 & 68.8 \\
\dashmid
\bf CPT \\
WD & - & 77.7 & 53.2 & 75.5 & 79.2 & 68.9 & 43.4 & 79.9 & 70.5 & 68.5 \\
no-WD & - & 77.6 & 53.3 & 75.4 & 79.1 & 69.6 & 43.6 & 80.0 & 70.6 & 68.7 \\
PN-D & 0.1 & 78.3 & 53.2 & 74.7 & 79.2 & 69.4 & 45.6 & 78.3 & 69.7 & 68.5 \\
PL-D & 0.1 & 79.0 & 54.3 & 74.9 & 79.4 & 69.1 & 44.6 & 81.0 & 70.4 & 69.1 \\
VPAD & 0.1 & 79.3 & 54.9 & 75.6 & 79.5 & 69.8 & 44.4 & 80.3 & 69.9 & 69.2 \\
PN-D & 0.2 & 78.3 & 53.0 & 73.4 & 79.2 & 67.7 & 45.0 & 78.5 & 69.8 & 68.1 \\
PL-D & 0.2 & 78.4 & 51.7 & 72.8 & 78.9 & 68.7 & 44.2 & 80.2 & 69.6 & 68.1 \\
VPAD & 0.2 & 79.4 & 54.4 & 74.1 & 79.9 & 68.3 & 44.2 & 80.2 & 69.2 & 68.7 \\
PN-D & 0.3 & 77.5 & 51.5 & 71.4 & 77.6 & 66.5 & 44.0 & 78.3 & 70.0 & 67.1 \\
PL-D & 0.3 & 76.6 & 47.7 & 68.1 & 77.1 & 67.2 & 43.0 & 77.7 & 67.8 & 65.7 \\
VPAD & 0.3 & 77.9 & 52.6 & 71.7 & 79.1 & 66.7 & 43.6 & 78.5 & 67.6 & 67.2 \\
\bottomrule
\end{tabular}
\end{table}

\subsection{Extended SmolLM3 Results on RULER}\label{app:smollm3-extended-ruler}
Table~\ref{tab:SmolLM3_ruler_extended} shows extended NIAH-1 and RULER results for various dropout probability values.

\begin{table}[ht]
\centering
\caption{SmolLM3 performance on NIAH-1 and on RULER for all dropout probabilities. Notation as in Table~\ref{tab:SmolLM3_niah_1}. ``--- $V$ ---'' means that all entries are equal to $V$ in the range; $V_1 \rightarrow V_2$ means that the left bound of the range has value $V_1$ and the right one $V_2$.
Grey columns --- ID length. }
\label{tab:SmolLM3_ruler_extended}
\resizebox{0.85\linewidth}{!}{%
\begin{tabular}{lrrrcrcrrrr}
\toprule
& \multicolumn{5}{c}{NIAH-1} & & \multicolumn{4}{c}{RULER} \\
\cmidrule(lr){2-6} \cmidrule(lr){8-11}
Model & \cellcolor{idgray}4K  & 8K & 16K & 32-128K & 256K & & \cellcolor{idgray}4K  & 8K & 16K & 32K \\
\midrule
Orig. & 100.0 & 0.0 & 0.0 &  --- 0.0 --- & 0.0 & & 83.5 & 0.1 & 0.0 & 0.0 \\
SW & 100.0 & 0.0 & 0.0 & --- 0.0 --- & 0.0 & & 83.4 & 0.5 & 0.0 & 0.0 \\
\dashmid
\bf CPT \\
SW, WD & 100.0 & 0.0 & 0.0 & --- 0.0 --- & 0.0 & & 83.5 & 1.1 & 0.0 & 0.0 \\
SW, no-WD & 100.0 & 0.0 & 0.0 & --- 0.0 ---  & 0.0 & & 83.5 & 1.1 & 0.0 & 0.0 \\
\dashmid
\bf CPT: PN-D \\
SW, PN-D $p=0.1$ &  100.0 & 0.0 & 0.0   &  --- 0.0 --- & 0.0 & & 80.8  & 2.6  & 0.7   & 0.2 \\
SW, PN-D $p=0.2$ &   100.0 & 0.6 & 0.0   &  --- 0.0 --- & 0.0 & & 78.4  & 3.7  & 1.9   & 1.5  \\
SW, PN-D $p=0.3$ &  100.0 & 82.2  & 6.2  & --- 0.0 --- & 0.0 & & 76.1 & 22.5 & 2.1  & 0.9 \\
\dashmid
\bf CPT: PL-D \\
SW, PL-D $p=0.1$ & 100.0 & 100.0 & 75.4  & $18.4 \rightarrow 0.2$ & 0.0 & &  79.7  & 47.8 & 13.7  & 3.4  \\
SW, PL-D $p=0.2$ &  100.0 & 100.0 & 100.0 & $100.0 \rightarrow 94.4$ & 36.6  & &  78.9  & 50.9 & 38.2  & 21.7 \\
SW, PL-D $p=0.3$ & 100.0 & 100.0 & 100.0 & --- 100.0 --- & 100.0 & & 74.1 & 50.6 & 40.2 & 25.5 \\
\dashmid
\multicolumn{3}{l}{\bf CPT: VPAD} \\
SW, VPAD $p=0.1$ & 100.0 & 100.0 & 57.2  & $30.4 \rightarrow 0.0$  & 0.0 & & 80.4 & 45.2 & 5.9 & 2.0 \\
SW, VPAD $p=0.2$ & 100.0 & 100.0 & 100.0 & $96.2 \rightarrow 86.0$  & 49.4 & & 79.9 & 55.0 & 38.8 & 19.4 \\
SW, VPAD $p=0.3$ & 100.0 & 100.0 & 100.0 &  --- 100.0 --- & 99.2 & & 77.8 & 54.3 & 44.1 & 27.4 \\
\bottomrule
\end{tabular}%
}
\end{table}

\subsection{Checkpoint Averaging and Attention Dropout}

\begin{figure}
    \includegraphics[width=0.33\linewidth]{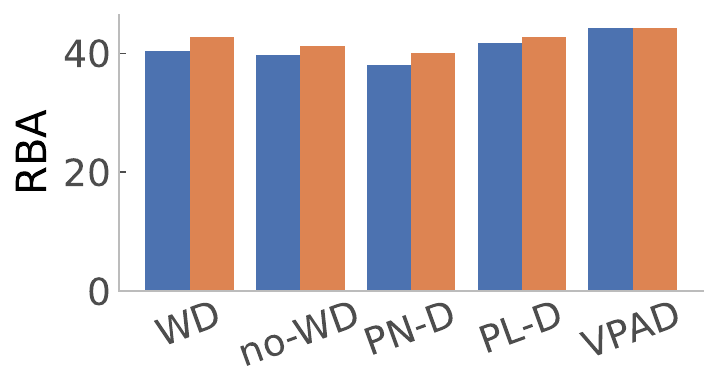}
    \includegraphics[width=0.33\linewidth]{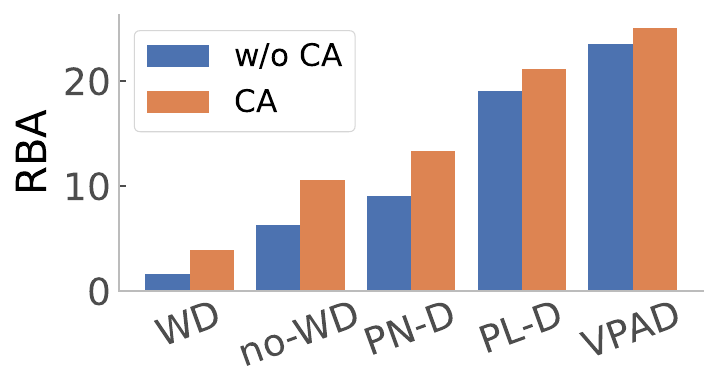}
    \includegraphics[width=0.33\linewidth]{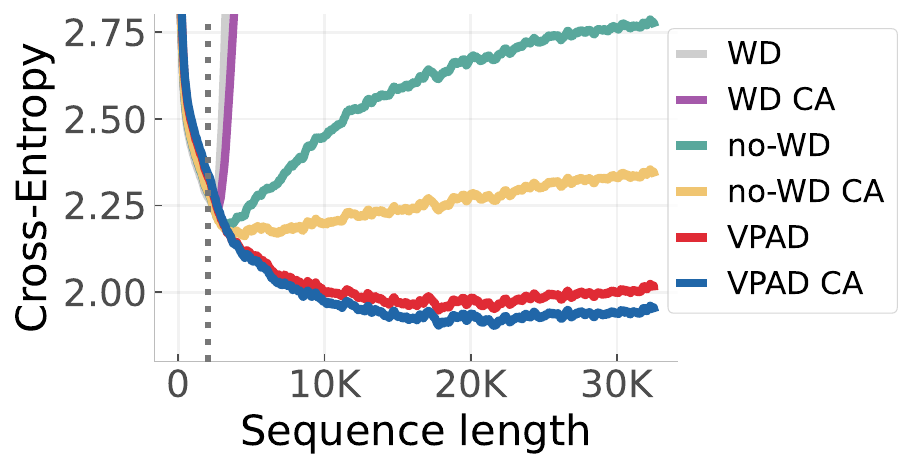}
     \caption{Performance of models trained from scratch before and after applying checkpoint averaging. \textbf{Left:} RULER: in-distribution length (2K); \textbf{Middle:} RULER: out-of-distribution---average of performance at 4K and 8K;
     \textbf{Right:} Cross-Entropy loss on ArXiv papers from The Common Pile.
     }
    \label{fig:checkpoint-avg}
\end{figure}

Checkpoint averaging (CA) builds a single model by averaging the weights of several checkpoints collected along one training trajectory, $\bar{\vv{\theta}} = \tfrac{1}{K}\sum_{t=1}^{K}\vv{\theta}_t$, where $\vv{\theta}_t$ are the parameters at the saved checkpoints. Because it reuses checkpoints already produced during training, CA improves generalization at essentially no additional cost, and it has long been a standard practice. For example, it is used in the original Transformer \citep{vaswani2017attention}. To some extent, CA can be related to stochastic weight averaging \citep{izmailov2018averaging} and, thus, can be considered as implicit regularization \citep{MoralesBrotons2024ExponentialMA}.
We believe that, by investigating regularization methods such as CA and dropout, we show that models suffer from \emph{length-based overfitting} that is not related to architectural components but rather is a consequence of the training process.

For the experiments with checkpoint averaging, we average 10 checkpoints of transformers, which constitute $25\%$ of the training span.
We compare the models trained from scratch before and after applying checkpoint averaging in Fig.~\ref{fig:checkpoint-avg} (see App.~\ref{appendix:experiment-details} for details). It is evident that the averaging improves the CE loss, especially for the no-WD model. Moreover, this improvement in CE loss transfers to RULER.
 We observe an improvement in the OOD performance of all models, including those with dropout (though with a diminishing effect), but the models trained without dropout benefit the most.
 The ID performance is slightly improved as well, except for VPAD, where the performance is unchanged.
In the case of SmolLM3, we observe that, even without CPT, CA is able to improve the extrapolation of the SW variant on NIAH-1 and on RULER overall (see Tables \ref{tab:attn-drop-niah-1} and \ref{tab:attn-drop-ruler}).
Note that the model was never exposed to a context length larger than 4096 and was never trained with a sliding window. However, it is capable of operating in an SW setting without any additional training.

Additionally, we observe extrapolation improvements when SmolLM3 is fine-tuned with attention dropout, which drops weights in the attention matrix (see Table~\ref{tab:attn-drop-ruler}). However, the behavior differs on NIAH-1, where attention dropout brings no improvement at all (Table~\ref{tab:attn-drop-niah-1}).

\begin{table}[ht]
\centering
\caption{Performance on NIAH-1 of SmolLM3 fine-tuned with attention dropout}
\label{tab:attn-drop-niah-1}
\begin{tabular}{lrrr}
\toprule
Model & 4K & 8K & 16K \\
\midrule
Orig. &  100.0 & 0.0 & 0.0 \\
\dashmid
\bf Checkpoint Avg. \\
SW, $K = 10$ &   100.0 & 41.2 & 0.8 \\
SW, $K = 20$ &  97.2  & 66.4  & 11.4  \\
SW, $K = 30$ &   80.4  & 68.2  & 2.2    \\
\dashmid
\bf CPT: Attention Dropout \\
SW, no-WD, $p=0.1$ & 100.0   & 0.0  & 0.0  \\
SW, no-WD, $p=0.2$ & 100.0   & 0.0 & 0.0   \\
SW, no-WD, $p=0.3$ & 100.0   & 0.0  & 0.0   \\

\bottomrule
\end{tabular}
\end{table}

\begin{table}[ht]
\centering
\caption{Performance on RULER of SmolLM3 fine-tuned with attention dropout}
\label{tab:attn-drop-ruler}
\begin{tabular}{lrrr}
\toprule
Model & 4K & 8K & 16K \\
\midrule
Orig. &  83.5 & 0.1 & 0.0 \\
\dashmid
\bf Checkpoint Avg. \\
SW, $K=10$ & 79.4  & 11.4 & 0.7   \\
SW, $K=20$ & 61.6  & 13.8 & 1.6  \\
SW, $K=30$ & 43.2  & 10.7 & 1.7   \\
\dashmid
\bf CPT: Attention Dropout \\
SW, no-WD, $p=0.1$ & 80.7  & 4.3  & 0.1  \\
SW, no-WD, $p=0.2$ & 76.2  & 13.4 & 0.9   \\
SW, no-WD, $p=0.3$ & 74.3  & 7.7  & 1.3   \\

\bottomrule
\end{tabular}
\end{table}

\subsection{ASEntmax}\label{app:asentmax-res}
\citet{vasylenko2026longcontext} showed that adaptive-sparsity transformers length-generalize better than dense (softmax-based) transformers. We observe a similar pattern, and dropout intensifies the effect (Table~\ref{tab:asentmax-ruler}).

\begin{table}[ht]
\centering
\caption{Performance of ASEntmax models on RULER}
\label{tab:asentmax-ruler}
\begin{tabular}{lrrrrr}
\toprule
Model & \cellcolor{idgray} 2K & 4K & 8K & 16K & 32K \\
\midrule
Softmax WD & 40.39	& 3.09 & 0.06 & 0.04 & 0.04 \\
Softmax VPAD & 44.24 & 34.76 & 12.30 & 4.54 & 1.93 \\
\dashmid
ASEntmax WD & 43.96 & 20.94 & 5.44 & 2.97 & 2.44 \\
ASEntmax PN-D & 42.76 & 26.85 & 7.72 & 2.36 & 1.59 \\
ASEntmax PL-D & 47.30 & 36.80 & 20.11 & 11.39 & 6.73 \\
ASEntmax VPAD & 48.09 & 37.32 & 21.64 & 13.51 & 6.98 \\

\bottomrule
\end{tabular}
\end{table}

\subsection{SmolLM3: NIAH-1 and Checkpoint Averaging}
Table~\ref{tab:SmolLM3_niah_1-full} shows extended NIAH-1 and checkpoint-averaging results.

\begin{table}[ht]
\centering
\caption{Full SmolLM3 results on NIAH-1.}
\label{tab:SmolLM3_niah_1-full}
\resizebox{1.0\linewidth}{!}{%
\begin{tabular}{lrrrrrrr}
\toprule
Model & \cellcolor{idgray} 4K & 8K & 16K & 32K & 64K & 128K & 256K \\
\midrule
SmolLM3 & 100.0 & 0.0 & 0.0 & 0.0 & 0.0 & 0.0 & 0.0 \\
SmolLM3-SW & 100.0 & 0.0 & 0.0 & 0.0 & 0.0 & 0.0 & 0.0 \\
\dashmid
\textbf{Checkpoint Averaging} \\

SmolLM3-SW, $K = 10$ &   100.0 & 41.2 & 0.8 & 0.0 & 0.0 & 0.0 & 0.0\\
SmolLM3-SW, $K = 20$ &  97.20  & 66.4  & 11.4 & 0.0 & 0.0 & 0.0 & 0.0    \\
SmolLM3-SW, $K = 30$ &   80.40  & 68.2  & 2.2 & 0.0 & 0.0 & 0.0 & 0.0    \\
\dashmid
\textbf{Continued Pretraining} \\
SmolLM3-SW, WD & 100.0 & 0.0 & 0.0 & 0.0 & 0.0 & 0.0 & 0.0\\
SmolLM3-SW, no-WD & 100.0 & 0.0 & 0.0 & 0.0 & 0.0 & 0.0 & 0.0  \\
\dashmid
SmolLM3-SW, PN-D $p=0.1$ & 100.0 & 0.0   & 0.0 & 0.0 & 0.0 & 0.0 & 0.0  \\
SmolLM3-SW, PN-D $p=0.2$ & 100.0 & 0.6   & 0.0   & 0.0 & 0.0 & 0.0 & 0.0  \\
SmolLM3-SW, PN-D $p=0.3$ & 100.0 & 82.2  & 6.2   & 0.2 & 0.0 & 0.0 & 0.0  \\
\dashmid
SmolLM3-SW, PL-D $p=0.1$ &  100.0 & 100.0 & 75.4  & 18.4  & 2.8   & 0.2    & 0.0  \\
SmolLM3-SW, PL-D $p=0.2$ & 100.0 & 100.0 & 100.0 & 100.0 & 100.0 & 94.4   & 36.6 \\
SmolLM3-SW, PL-D $p=0.3$ & 100.0 & 100.0 & 100.0 & 100.0 & 100.0 & 100.0 & 100.0 \\
\dashmid
SmolLM3-SW, VPAD $p=0.1$ &  100.0 & 100.0 & 57.2  & 30.4  & 5.6  & 0.0 & 0.0 \\
SmolLM3-SW, VPAD $p=0.2$ & 100.0 & 100.0 & 100.0 & 96.2  & 91.6  & 86.0   & 49.4  \\
SmolLM3-SW, VPAD $p=0.3$ & 100.0 & 100.0 & 100.0 & 100.0 & 100.0 & 100.0 & 99.20 \\
\bottomrule
\end{tabular}
}
\end{table}

\subsection{Mamba2}\label{app:mamba-std}
Tables~\ref{tab:res-mamba-copy-std} and~\ref{tab:res-mamba-85m-niah-1-std} show the results of the small models trained from scratch on synthetic copying and NIAH-1. Table~\ref{tab:res-mamba-ruler-full} shows extended results on NIAH-1 and RULER.

\begin{table}[ht]
\centering
\begin{minipage}[t]{0.38\linewidth}
    \centering
    \caption{Extrapolation results of Mamba2 models on the synthetic copying task (mean $\pm$ standard deviation over 5 seeds)}
    \label{tab:res-mamba-copy-std}
    \resizebox{\linewidth}{!}{%
    \begin{tabular}{lll}
        \toprule
        Model & EMA-150 & CLA-300  \\
        \midrule
        WD & $33.1\,{\scriptstyle\pm\,21.9}$ & $42.8\,{\scriptstyle\pm\,12.7}$ \\
        no-WD & $42.7\,{\scriptstyle\pm\,13.7}$ & $45.7\,{\scriptstyle\pm\,14.5}$\\
        PL-D & $61.5\,{\scriptstyle\pm\,19.6}$ & $50.9\,{\scriptstyle\pm\,8.9}$\\
        VPAD & $68.9\,{\scriptstyle\pm\,13.6}$ & $58.0\,{\scriptstyle\pm\,10.8}$\\
        \bottomrule
    \end{tabular}
    }
\end{minipage}
\hfill
\begin{minipage}[t]{0.56\linewidth}
    \centering
    \caption{Mamba2-85M performance on NIAH-1 (mean $\pm$ standard deviation over 3 seeds)}
    \label{tab:res-mamba-85m-niah-1-std}
    \resizebox{\linewidth}{!}{%
    \begin{tabular}{llll}
        \toprule
        Model & \cellcolor{idgray} 1K & 2K & 4K  \\
        \midrule
        WD &  $72.3\,{\scriptstyle\pm\,17.8}$ & $30.5\,{\scriptstyle\pm\,16.1}$  & $1.1\,{\scriptstyle\pm\,1.1}$ \\
        no-WD & $91.2\,{\scriptstyle\pm\,11.1}$ & $50.6\,{\scriptstyle\pm\,31.2}$  & $1.1\,{\scriptstyle\pm\,0.8}$ \\
        PL-D & $90.3\,{\scriptstyle\pm\,4.4}$ & $43.4\,{\scriptstyle\pm\,19.5} $ & $10.5\,{\scriptstyle\pm\,11.1}$ \\
        VPAD & $88.9\,{\scriptstyle\pm\,13.7}$ & $76.4\,{\scriptstyle\pm\,20.3}$  & $30.2\,{\scriptstyle\pm\,22.6}$ \\
        \bottomrule
    \end{tabular}
    }
\end{minipage}
\end{table}

\begin{table}[ht!]
\centering
\caption{Mamba2-1.3B performance on NIAH-1 and on RULER.
}
\label{tab:res-mamba-ruler-full}
\resizebox{0.88\linewidth}{!}{%
\begin{tabular}{lrrrrrcrrrrr}
\toprule
& \multicolumn{5}{c}{NIAH-1} & & \multicolumn{5}{c}{RULER} \\
\cmidrule(lr){2-6} \cmidrule(lr){8-12}
Model & \cellcolor{idgray}2K  & 4K & 8K & 16K & 32K & & \cellcolor{idgray}2K  & 4K & 8K & 16K & 32K \\
\midrule
Orig. & 100.0 & 100.0 & 100.0 & 0.0 & 0.0 & & 38.3 & 14.3 & 8.5  & 0.4  & 0.0 \\
\dashmid
\bf CPT \\
WD & 100.0 & 100.0 & 100.0 & 0.0 & 0.0 & & 42.0 & 16.2 & 9.6  & 0.7   & 0.0 \\
no-WD & 100.0 & 100.0 & 100.0 & 0.0 & 0.0 & & 42.2 & 16.0 & 9.6  & 0.0   & 0.0 \\
\dashmid
PN-D $p=0.2$ & 100.0 & 100.0 & 100.0 & 11.8 & 0.0 & & 50.7 & 22.5 & 11.5 & 2.6 & 0.8 \\
PN-D $p=0.3$ & 100.0 & 100.0 & 99.8 & 39.0 & 0.4 & & 50.9 & 24.8 & 12.8 & 5.1 & 1.7 \\
\dashmid
PL-D $p=0.1$ & 99.8  & 100.0 & 100.0 & 99.8  & 7.0 & & 48.9 & 23.0 & 14.7 & 10.5  & 1.5 \\
PL-D $p=0.2$ &  100.0 & 100.0 & 100.0 & 100.0 & 22.8 & & 48.9 & 28.8 & 18.0 & 13.7  & 5.3 \\
PL-D $p=0.3$ &  100.0 & 100.0 & 100.0 & 97.2  & 12.0 & & 46.5 & 31.4 & 20.0 & 13.9  & 4.8 \\
\dashmid
VPAD $p=0.1$ & 100.0 & 100.0 & 100.0 & 99.8  & 13.0 & & 50.6 & 23.1 & 14.8 & 10.4  & 2.0 \\
VPAD $p=0.2$ &  100.0 & 100.0 & 100.0 & 100.0 & 45.0 & & 50.9 & 26.3 & 18.5 & 13.3  & 5.8 \\
VPAD $p=0.3$ & 100.0 & 100.0 & 100.0 & 99.8  & 23.8 & & 46.6 & 24.8 & 18.9 & 15.7  & 5.5 \\
\bottomrule
\end{tabular}%
}
\end{table}

\end{document}